\documentclass{article}
\usepackage{arxiv}

\usepackage[numbers]{natbib}
\usepackage{url}
\usepackage{tikz}
\usepackage{pgfplots}
\usepackage{pgfplotstable}
\usepackage{color}
\usepackage{amsmath}
\usepackage{booktabs}
\usepackage{mathtools}
\usepackage[linesnumbered,algoruled,boxed,lined]{algorithm2e}
\usepackage{subcaption}
\usepackage{caption}
\usepackage{varwidth}
\newsavebox\tmpbox
\usepackage[noend]{algpseudocode}
\usepackage{amssymb}% http://ctan.org/pkg/amssymb
\usepackage{pifont}% http://ctan.org/pkg/pifont
\usepackage{fontawesome}
\usepackage[inline]{enumitem}
\usepackage{dblfloatfix}
\usepackage{diagbox}
\usepackage{bbm}
\newcommand{\cmark}{\ding{51}}%
\newcommand{\xmark}{\ding{55}}%
\usetikzlibrary{calc,positioning,shapes.geometric,arrows,fit}

\newcommand*\circled[1]{\tikz[baseline=(char.base)]{
            \node[shape=circle,draw,inner sep=1.2pt,fill=gray!10] (char) {#1};}}

\makeatletter
\tikzset{
    database/.style={
        path picture={
            \draw (0, 1.5*\database@segmentheight) circle [x radius=\database@radius,y radius=\database@aspectratio*\database@radius];
            \draw (-\database@radius, 0.5*\database@segmentheight) arc [start angle=180,end angle=360,x radius=\database@radius, y radius=\database@aspectratio*\database@radius];
            \draw (-\database@radius,-0.5*\database@segmentheight) arc [start angle=180,end angle=360,x radius=\database@radius, y radius=\database@aspectratio*\database@radius];
            \draw (-\database@radius,1.5*\database@segmentheight) -- ++(0,-3*\database@segmentheight) arc [start angle=180,end angle=360,x radius=\database@radius, y radius=\database@aspectratio*\database@radius] -- ++(0,3*\database@segmentheight);
        },
        minimum width=2*\database@radius + \pgflinewidth,
        minimum height=3*\database@segmentheight + 2*\database@aspectratio*\database@radius + \pgflinewidth,
    },
    database segment height/.store in=\database@segmentheight,
    database radius/.store in=\database@radius,
    database aspect ratio/.store in=\database@aspectratio,
    database segment height=0.1cm,
    database radius=0.25cm,
    database aspect ratio=0.35,
}
\makeatother

\DeclareMathOperator{\range}{range}

\newcommand{\todo}[1]{}
\renewcommand{\todo}[1]{{\color{red}{#1}}}
\newcommand{\todov}[1]{}
\renewcommand{\todov}[1]{{#1}}
\newcommand{\clearcolor}[1]{}
\renewcommand{\clearcolor}[1]{{\color{black}{#1}}}
\newcommand{\todoa}[1]{}
\renewcommand{\todoa}[1]{{#1}}
\newcommand{\todos}[1]{}
\renewcommand{\todos}[1]{{#1}}
\newcommand{\todoy}[1]{}
\renewcommand{\todoy}[1]{{#1}}
\newcommand{\shortciteauthor}[1]{}
\renewcommand{\shortciteauthor}[1]{\citet{#1}}

\usepackage{filecontents}

\newcommand{\plotfile}[1]{
    \pgfplotstableread{#1}{\table}
    \pgfplotstablegetcolsof{#1}
    \pgfmathtruncatemacro\numberofcols{\pgfplotsretval-1}
    \pgfplotsinvokeforeach{1,...,\numberofcols}{
        \pgfplotstablegetcolumnnamebyindex{##1}\of{\table}\to{\colname}
        \addplot table [y index=##1] {#1}; 
        \addlegendentryexpanded{\colname}
    }
}

\usepackage[utf8]{inputenc} % allow utf-8 input
\usepackage[T1]{fontenc}    % use 8-bit T1 fonts
\usepackage{hyperref}       % hyperlinks
\usepackage{url}            % simple URL typesetting
\usepackage{booktabs}       % professional-quality tables
\usepackage{amsfonts}       % blackboard math symbols
\usepackage{nicefrac}       % compact symbols for 1/2, etc.
\usepackage{microtype}      % microtypography
\usepackage{lipsum}		% Can be removed after putting your text content

\title{A framework for the extraction of Deep Neural Networks by leveraging public data}

\date{} 					% Or removing it

\author{
  Soham Pal\textsuperscript{1}\thanks{All three authors contributed equally.}, Yash Gupta\textsuperscript{1}\footnotemark[1], Aditya Shukla\textsuperscript{1}\footnotemark[1], Aditya Kanade\textsuperscript{1,2} \thanks{All three authors contributed equally.}, Shirish Shevade\textsuperscript{1}\footnotemark[2], Vinod Ganapathy\textsuperscript{1}\footnotemark[2] \medskip \\
\footnotesize
\textsuperscript{1}  Department of Computer Science and Automation, IISc Bangalore, India\\
\footnotesize
\textsuperscript{2} Google Brain, USA \medskip \\
\footnotesize
  \texttt{\{sohampal,yashgupta,adityashukla,kanade,shirish,vg\}@iisc.ac.in} \\
}

\begin{document}
\maketitle

\begin{abstract}
Machine learning models trained on confidential datasets are increasingly being deployed for profit. Machine Learning as a Service (MLaaS) has made such models easily accessible to end-users. Prior work has developed model extraction attacks, in which an adversary extracts an approximation of MLaaS models by making black-box queries to it. However, none of these works is able to satisfy \todoy{all the three essential criteria for practical model extraction}: \todov{\begin{enumerate*}[label=(\roman*)]\item the ability to work on deep learning models, \item the non-requirement of domain knowledge and \item the ability to work with a limited \todoy{query} budget\end{enumerate*}}. We design a model extraction framework that makes use of active learning and large public datasets to satisfy them. We demonstrate that it is possible to use this framework to steal deep classifiers trained on a variety of datasets from image and text domains. By querying a model via black-box access for its top prediction, \todos{our framework improves performance on an average over a uniform noise baseline by $\mathbf{4.70\times}$ for image tasks and $\mathbf{2.11\times}$ for text tasks respectively, while using only 30\% (30,000 samples) of the public dataset at its disposal.}
\end{abstract}

% keywords can be removed

\section{Introduction}
\label{sec:introduction}
Due to their success in recent years, deep neural networks (DNNs) are increasingly being deployed in production software. The security of these models is thus of paramount importance. The most common attacks against DNNs focus on the generation of adversarial examples \cite{goodfellow2014explaining,DBLP:journals/corr/abs-1712-07107,carlini2017towards,su2019one,papernot2016limitations,moosavi2016deepfool,DBLP:journals/corr/PapernotMGJCS16,nguyen2015deep}, where attackers add an imperceptible perturbation to inputs (typically images) that cause DNNs to misclassify them.

In this paper, we turn our attention to privacy vulnerabilities. \todos{Today, Machine Learning as a Service (MLaaS) providers like Google, Amazon and Azure make ML models available through APIs to developers of web and mobile applications}. These services are monetized by billing queries pro rata. The business model of these services rests on the privacy of the model. If it was possible for a potential competitor or end user to create a copy of these models with access only to the query API, \todoy{it would pose a great threat to their business}.

By extracting a copy of a ML model, not only would an adversary have the ability to make unlimited free queries to it, they would also be able to implement applications requiring gradient information, such as crafting adversarial examples that fool the secret MLaaS model \cite{DBLP:journals/corr/PapernotMGJCS16}, performing model inversion \cite{fredrikson2015model} (discovering the training data on which the model was originally trained) and exploring the explainability of proprietary ML models (e.g., by training an explainable substitute model such as a decision tree classifier \cite{bastani2017interpretability}).

Model privacy is also important to developers of other ML products (such as self-driving vehicles and translation tools). Datasets are expensive to gather and curate, and models require expertise to design and implement -- thus, it is in the best interest of corporations to protect their ML models to maintain a competitive edge.

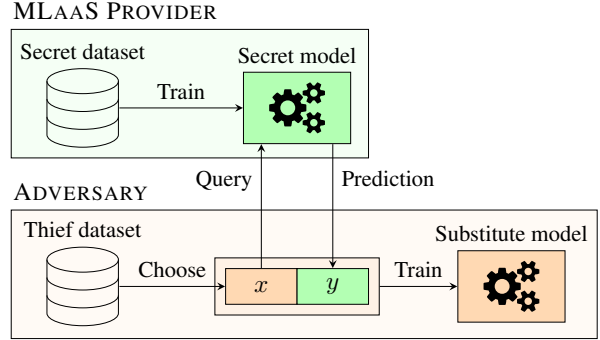
\begin{figure}[t!]
\centering
\scalebox{.95}{
\begin{tikzpicture}[
    >=stealth,
    node distance=3cm,
    block/.style={
      rectangle,
      aspect=0.25,
      draw,
      align=center
    }
  ]
  
 \node[block,minimum width=5cm,minimum height=1.8cm,fill=green!4] (encloseSecret) at (1.5,2.2) {};
 \node[block,minimum width=8.2cm,minimum height=1.8cm,fill=orange!4] (encloseThief) at (3.1,-0.33) {};
   
  \node[database,label={[align=center]above:\footnotesize Secret dataset},database radius=0.5cm,database segment height=0.25cm] (secret-dataset) at (0,2) {};
  \node[database,label={[align=center]above:\footnotesize Thief dataset},database radius=0.5cm,database segment height=0.25cm] (thief-dataset) at (0,-0.5) {};

  \node[block,minimum width=1.5cm,minimum height=1.0cm,fill=green!30,label={[align=center]above:\footnotesize Secret model}] (secret-model) at (3,2) {\huge \faGears};
  \node[block,minimum width=1.5cm,minimum height=1.0cm,fill=orange!30,label={[align=center]above:\footnotesize Substitute model}] (substitute-model) at (6,-0.5) {\huge \faGears};

 \node[block,minimum width=2.3cm,minimum height=0.8cm,fill=orange!10] (encloseX-Y) at (3.0,-0.5) {};
 \node[block,minimum width=1cm,minimum height=0.5cm,fill=orange!30] (X) at (2.5,-0.5) {$x$};
 \node[block,minimum width=1cm,minimum height=0.5cm,fill=green!30] (Y) at (3.5,-0.5) {$y$};

   \path[->,draw] (secret-dataset) edge node[above] {\footnotesize Train} (secret-model);
   \path[->,draw] (encloseX-Y) edge node[above] {\footnotesize Train} (substitute-model);
  
   \path[->,draw] (thief-dataset) edge node[above] {\footnotesize Choose} (X);
 
   \path[->,draw] (X.north) to node[align=center, left] {\footnotesize Query \\\\} (X.north |- secret-model.south);
   
   \path[->,draw] (Y.north |- secret-model.south) to node[align=center, right] {\footnotesize Prediction \\\\} (Y.north);
   
  \node[anchor=west,align=left] (above_noise) at (-1.1,0.83) {\textsc{Adversary}};
  \node[anchor=west,align=left] (above_noise) at (-1.1,3.36) {\textsc{MLaaS Provider}};
  
\end{tikzpicture}}
\caption{Overview of model extraction}
\label{fig:modelex}
\end{figure}

\todos{\shortciteauthor{DBLP:journals/corr/TramerZJRR16}} define the concept of \textit{model extraction} \todov{(see Figure~\ref{fig:modelex})}. In model extraction, the adversary is an agent that can query a \textbf{secret model} (e.g., \todos{a MLaaS provider} via APIs) to obtain predictions on any supplied input vector of its choosing. \todos{The returned predictions may either be label probability distributions, or just the Top-1 prediction -- we assume the latter. Using the obtained predictions, the adversary trains a \textbf{substitute model} to approximate the secret model function.} The adversary may not know the secret model architecture or associated hyperparameters. \todos{The adversary has access to a \textbf{thief dataset} of the same media type (i.e.\ images or text) from which it draws samples to query the secret model. The data in this thief dataset may be drawn from a different distribution than the \textbf{secret dataset} on which the secret model was originally trained. Prior work has used the following thief datasets:}

\begin{itemize}
\item \textit{\clearcolor{Uniform noise}}: \todos{\shortciteauthor{DBLP:journals/corr/TramerZJRR16}} perform model extraction by querying the secret model with inputs sampled \textit{i.i.d.}\ uniformly at random. They demonstrate their method on logistic regression models, SVMs, shallow (1 hidden layer) feedforward neural networks and decision trees. According to our experiments, this approach does not scale well to deeper neural networks \todoy{(such as our architecture for image classification with 12 convolutional layers; see Section~\ref{sec:image-results} for further details)}.

\item \textit{\clearcolor{Hand-crafted examples}}: \todos{\shortciteauthor{DBLP:journals/corr/PapernotMGJCS16}} design a model extraction framework that can be used to extract DNNs. However, this technique assumes domain knowledge on the part of the attacker. The adversary should either have access to a subset of the secret dataset, or create data (such as by drawing digits using a pen tablet) that \todoy{closely} resembles it.

\item \textit{\clearcolor{Unlabeled non-problem domain data}}: \todos{\shortciteauthor{DBLP:journals/corr/abs-1806-05476}} demonstrate that convolutional neural networks (CNNs) can be copied by querying them with a mix of non-problem domain and problem domain data. For example, they demonstrate that a DNN trained using European crosswalk images \cite{7979607} as the secret dataset can be copied using a mix of ImageNet (non-problem domain data) and crosswalk images from America and Asia (problem domain data) as the thief dataset. They do not consider a query budget in their work.
\end{itemize}

In this work, we investigate the feasibility of implementing a practical approach to model extraction, viz.\ one that deals with the following criteria:

\begin{itemize}
\item \textbf{Ability to extract DNNs:} \todoy{Most state of the art ML solutions use DNNs.} Thus, it is critical for a model extraction technique to be effective for this class of models.
\item \textbf{No domain knowledge:} \todoy{The adversary should be expected to have little to no domain knowledge related to task implemented by the secret model.} In particular, they should not be expected to have access to samples from the secret dataset.
\item \textbf{Ability to work within a query budget:} \todoy{Queries made to MLaaS services are billed pro rata, and such services are often rate limited.} Thus, it is in an attacker's best interest to minimize the number of queries they make to the secret model.
\end{itemize}

\begin{table}[t!]
\centering
\caption{Comparison of Model Extraction approaches}
\begin{tabular}{lccc}  
\toprule
Model extraction & Works on & No domain & Limited \# \\
technique  & DNNs & knowledge & of queries \\
\midrule
Tram{\`e}r et al. \cite{DBLP:journals/corr/TramerZJRR16}  & \color{red}{\xmark} & \cmark & \cmark \\
Papernot et al. \cite{DBLP:journals/corr/PapernotMGJCS16} & \cmark & \color{red}{\xmark} & \cmark \\
Copycat CNN \cite{DBLP:journals/corr/abs-1806-05476} & \cmark & \cmark & \color{red}{\xmark}  \\
\textbf{Our framework}  & \cmark & \cmark & \cmark  \\
\bottomrule
\end{tabular}
\label{tab:comparison}
\end{table}

We compare our approach to the three approaches described above on these three criteria \cite{DBLP:journals/corr/TramerZJRR16,DBLP:journals/corr/PapernotMGJCS16,DBLP:journals/corr/abs-1806-05476} in Table~\ref{tab:comparison}. As can be seen, we can extract DNNs with no domain knowledge, while working with a limited query budget. To achieve these criteria, our paper introduces two novel techniques:

\begin{itemize}
\item \textbf{Universal thief datasets:} These are large and diverse public domain datasets, analogous to the non-problem domain (NPD) data of \shortciteauthor{DBLP:journals/corr/abs-1806-05476}. For instance, we show that ImageNet constitutes a universal thief for vision tasks, whereas a dataset of Wikipedia articles constitutes a universal thief for NLP tasks. Our key insight is that universal thief datasets provide a more natural prior than uniform noise, while not requiring domain knowledge to obtain.

\item \textbf{Active learning strategies:} Active learning is a technique used in scenarios where labeling is expensive. It strives to select a small yet informative set of training samples to maximize accuracy while minimizing the total labeling cost. In this paper, we use \textit{pool-based active learning}, where the algorithm has access to a large set of  unlabeled examples (i.e.\ the thief dataset) from which it picks the next sample(s) to be labeled.

Although universal thief datasets constitute an excellent prior for model extraction, their size makes them unsuitable for use when the query budget is limited. We make use of active learning to construct an optimal query set, thus reducing the number of queries made to the MLaaS model. This ensures that the attacker stays within the query budget.
\end{itemize}

Our contributions include:

\begin{enumerate}
\item We define the notion of \todos{universal thief datasets for different media types such as images and text}.
\item We propose a framework for model extraction that makes use of universal thief datasets in conjunction with active learning strategies. We demonstrate our framework on DNNs for image \todov{and text} classification tasks.
\item Finally, we introduce the notion of \textit{ensemble active learning strategies} as a combination of existing active learning strategies. We design and leverage one such ensemble strategy to improve performance.
\end{enumerate}

Overall, we demonstrate that by leveraging public data and active learning, we improve agreement between the secret model and the substitute model by, on an average, $\mathbf{4.70\times}$ (across image classification tasks) and $\mathbf{2.11\times}$ (across text classification tasks) over the uniform noise baseline of \shortciteauthor{DBLP:journals/corr/TramerZJRR16}, when working with a total query budget of 30K.

We plan to release the source code for our framework under an open source license soon.

\section{Background}
\label{sec:background}

In this section, we introduce the \textit{active learning} set of techniques from the machine learning literature. We also briefly discuss adversarial example generation, which is later used as the crux of the DeepFool Active Learning (DFAL) strategy \cite{DBLP:journals/corr/abs-1802-09841} used by our framework.

\subsection{Preliminaries}

In machine learning, a \textit{dataset} $\mathcal{D}$ consists of \textit{labeled examples} $(x, y)$, where $x \in \mathcal{X}$ is an \textit{example} and $y \in \mathcal{Y}$ is its associated \textit{label}, where $\mathcal{X}$ is said to be the \textit{instance space} and $\mathcal{Y}$ is the \textit{label space}. It is assumed that there is an underlying unknown mapping $\phi: \mathcal{X} \to \mathcal{Y}$ from which $\mathcal{D}$ is generated (i.e.\ $(x, y) \in \mathcal{D}$ implies that $y = \phi(x)$). In this paper, we restrict ourselves to the classification setting, where $\mathcal{Y} = \{e_1, e_2, \dots, e_J\}$\footnote{$e_j$ represents the $j$\textsuperscript{th} standard basis vector, i.e.\ $\langle 0, 0, \dots, 0, 1, 0, \dots, 0, 0 \rangle \in \mathbb{R}^J$, a vector with a a $1$ in the $j$\textsuperscript{th} position, and $0$ elsewhere. Such a vector is said to be \textbf{one-hot}. A pair $(x,y)$ where $y$ is a vector with $1$ in the $j$\textsuperscript{th} position indicates that the sample $x$ belongs to the $j$\textsuperscript{th} class (out of $J$ classes).}.

In passive machine learning, the learner has access to a large \textit{training dataset} $\mathcal{D}_\text{train}$ of labeled examples and must learn a \textit{hypothesis function} $f$ that minimizes a \textit{loss function}. A typical loss function is mean squared error (MSE):
\begin{equation*}
\mathcal{L}_\text{MSE}(f, \mathcal{D}_\text{train}) = \frac{1}{\left|\mathcal{D}_\text{train}\right|} \sum_{(x,y) \in \mathcal{D}_\text{train}} \lVert y - f(x) \rVert^2_2
\end{equation*}

The better the hypothesis (i.e.\ when predictions $f(x)$ match labels $y$), the lower the value of the loss function $\mathcal{L}$. Other loss functions such as cross-entropy (CE) are also used. Machine learning models such as DNNs learn a function by minimizing this loss function on the training dataset. DNNs, when trained on a large corpus of training examples, have been shown to exhibit good \textit{generalization} ability across a diversity of tasks in various domains \cite{lecun2015deep}, i.e.\ provided a previously unseen \textit{test example} $x_\text{test}$, the prediction that they make, $f(x_\text{test})$ approximates the value of $\phi(x_\text{test})$ well, i.e.\ $f(x_\text{test}) \approx \phi(x_\text{test})$.

However, to achieve good generalization performance, such DNNs require a very large training dataset. The labeling effort required is massive, and learning may be intractable in scenarios where there is a high cost associated with each label, such as paying crowd workers. In the context of model extraction, this may involve querying a MLaaS model, which are billed pro rata by the MLaaS service provider.

\subsection{Active learning}
Active learning \cite{settles2009active} is useful in scenarios where there is a high cost associated with labeling instances. In active learning, the learner does not use the full labeled dataset $\mathcal{D}$. Rather, the learner starts with either an unlabeled dataset $X$ of samples $x$; or, alternatively, the learner can itself generate samples $x$ de novo. Following this, an oracle $f_\mathcal{O}$ is used to label the sample, which assigns it the true label $y = f_\mathcal{O}(x)$. Active learning can be broadly classified into one of the following scenarios:

\begin{itemize}
\item \textit{Stream-based selective sampling}:
In this scenario, the learner is presented with a stream of unlabeled samples $x_1, x_2, x_3, \dots$, drawn from the underlying distribution. The learner must decide to either accept or reject an individual sample $x_n$ for querying. This can be done by checking, e.g., the ``uncertainty'' of the prediction (we will formally define this in Section~\ref{sec:strategies}) made by the classifier on a specific sample $x_n$. For samples that are accepted by the learner, the oracle is queried to label them. Once rejected, a sample cannot be queried in the future.

\item \textit{Pool-based sampling}:
In this scenario, the learner has access to a full unlabeled dataset $X$ of samples $\{x_1, x_2, \dots x_{\left| X \right|}\}$. Unlike in stream-based selective sampling, the learner does not have to consider each sample $x_n$ in isolation. The learner's objective is thus to select a subset $S \subseteq X$ of samples to be queried. While it is possible to do this in one shot, pool-based sampling may also be done incrementally, either choosing one sample at a time, or an entire batch of samples in each iteration. Correspondingly, the oracle may be queried on one sample at a time, or the entire batch of selected samples.

\item \textit{Query synthesis}:
Here, the learner generates samples $x$ de novo without first approximating the underlying distribution. This process could be entirely uninformed -- for instance, the learner could generate data points by sampling uniformly at random from a multivariate uniform or Gaussian distribution -- or, it could be more informed: such as by using a generative model. The oracle is then queried with the generated sample.
\end{itemize}

In this work, we make use of \textit{pool-based sampling}. In particular, we consider the scenario where the learner adds a batch of samples in each iteration of the algorithm. We grow the subset $S_0 \subsetneq S_1 \subsetneq S_2 \subsetneq \cdots \subsetneq S_N$ over $N$ iterations, such that each subset $S_i$ is a selection of samples from the full dataset $S_i \subseteq X$.

\subsection{Adversarial example generation and the DeepFool technique}
\label{sec:deepfool}

We introduce the notion of adversarial example generation, in particular the DeepFool \cite{moosavi2016deepfool} technique. This technique will be used while introducing the DeepFool Active Learning (DFAL) \cite{DBLP:journals/corr/abs-1802-09841} active learning strategy in Section~\ref{sec:strategies}.

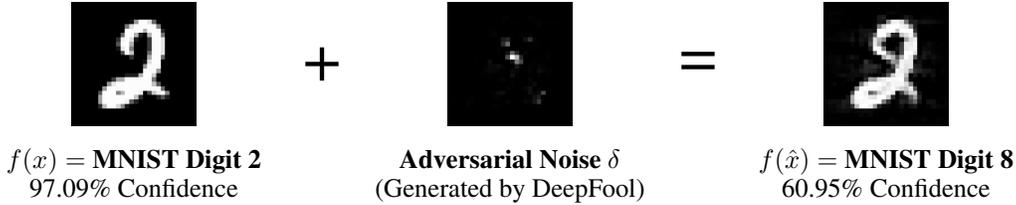
\begin{figure*}[t!]
\centering
\begin{tikzpicture}
\node[inner sep=0pt] at (0,0) {\includegraphics[width=.1\textwidth]{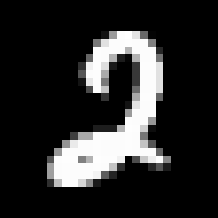}};
\node[inner sep=0pt] at (0,-1.5) {\begin{tabular}{c}$f(x) =$ \textbf{MNIST Digit 2} \\ 97.09\% Confidence\end{tabular}};
\node[inner sep=0pt] at (2.5, 0) {\Huge +}; 
\node[inner sep=0pt] at (5,0) {\includegraphics[width=.1\textwidth]{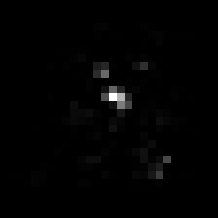}};
\node[inner sep=0pt] at (5,-1.5) {\begin{tabular}{c}\textbf{Adversarial Noise} $\delta$ \\ (Generated by DeepFool)\end{tabular}}; 
\node[inner sep=0pt] at (7.5, 0) {\Huge =}; 
\node[inner sep=0pt] at (10,0) {\includegraphics[width=.1\textwidth]{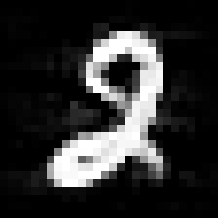}};
\node[inner sep=0pt] at (10,-1.5) {\begin{tabular}{c}$f(\hat x) =$ \textbf{MNIST Digit 8} \\ 60.95\% Confidence\end{tabular}}; 
%\draw[<->,thick] (russell.south east) -- (whitehead.north west) node[midway,fill=white] {Principia Mathematica};
\end{tikzpicture}

\caption{Adversarial example generation using DeepFool \cite{moosavi2016deepfool}.}
\label{fig:deepfool}
\end{figure*}

It is known that DNNs can be easily fooled as demonstrated by, the Fast Gradient Sign Method (FGSM) of \shortciteauthor{goodfellow2014explaining}, the C\&W attack of \shortciteauthor{carlini2017towards}, the Jacobian-based Saliency Map Attack (JSMA) of \shortciteauthor{papernot2016limitations} and many others \cite{DBLP:journals/corr/abs-1712-07107,su2019one,moosavi2016deepfool,DBLP:journals/corr/PapernotMGJCS16,nguyen2015deep}. In particular, neural networks trained to perform image classification tasks have been shown to be vulnerable to adversarial examples. An adversary can add a small amount of noise to input images, which, while being imperceptible to the human eye, can change the classification decision made by the neural network, as shown in Figure~\ref{fig:deepfool}.

These techniques typically work as follows -- given an innocuous image $x$, they compute a small, typically imperceptible additive noise $\delta$. This noise is then added to the original image to produce an adversarial image, $\hat x = x + \delta$. The objective is that, given a machine learning model $f$, the prediction of the perturbed image no longer matches the prediction made for the original image, viz.\ $f(x) \neq f(\hat x)$.

DeepFool \cite{moosavi2016deepfool} is one such technique for the generation of adversarial examples. It solves the following problem iteratively:
\begin{equation*}
\delta^* = \arg\min_\delta \lVert \delta \rVert_2 \text{ s.t. } f(x + \delta) \neq f(x)
\end{equation*}
In the binary classification setting (i.e.\ where $\range f = \{-1, 1\}$), it uses a first order approximation of the analytical solution for the linearly-separable case:
\begin{align*}
\delta_l &= - \frac{f(x_l)}{\lVert \nabla f(x_l) \rVert^2_2} \nabla f(x_l)\\
x_{l+1} &= x_l + \delta_l
\end{align*}
The process is started by setting $x_0 = x$, and terminates at the lowest index $L$ for which $f(x_L) \neq f(x)$. The total perturbation is obtained by taking the sum of the individual perturbations at each step, $\delta = \sum_{l=1}^L \delta_l$. This algorithm can be extended to work in the multiclass classification setting. We refer interested readers to \cite{nguyen2015deep} for further details.

\section{Threat model}

Before we describe the proposed algorithm, we first state the threat model under which it operates. 

\paragraph{Attack surface.}
We assume that the adversary cannot directly access the secret model, but can only query it in a black-box fashion via an API. We assume that there is a \textit{query cost} associated with each API query made by the adversary. While there is no limit on the number of queries that can be made theoretically, the ability of the adversary to make queries is restricted in practice by the \textit{total query budget}. This query cost model can be used to model rate-limiting defenses. For example, each query can have an associated cost, and a defense would be to limit queries from a source that has exceeded its cost threshold.

\paragraph{Capabilities.}
The adversary has black-box access to the {secret model} via an API, by which it can query it with any image or text of its choosing. It thus has full knowledge of both the input specification (i.e.\ the type of media -- images or text) and the output specification (the set of possible labels). Note that the adversary does not have direct access to the exact gradient information of the model, but only the final prediction. We consider two scenarios -- one where a Top-1 prediction is returned (as a one-hot standard basis vector), and another where the model returns a softmax\footnote{Given unnormalized scores $a_1, a_2, \dots a_J$ over $J$ classes, the softmax function computes the normalized quantities $p_i = \exp(a_i)/\sum_{j=1}^J \exp(a_j)$. The resulting $p_i$ values constitute a valid probability distribution.} probability distribution over the target output classes. Our primary experiments assume the weaker capability of receiving only the Top-1 predictions, and not the softmax probability distributions.\footnote{However, in Table \ref{tab:num_iter_softmax} we also consider the situation where the softmax probability distribution is available to the adversary.}

Information of the secret model architecture and model hyperparameters need not be known to the adversary, as we show in Section~\ref{sec:grids}. However, as peak performance is achieved when the adversary is aware of the architecture of the secret model, and since it is possible to detect these hyperparameters and architectural choices by a related line of work (\textit{model reverse-engineering} \cite{2017arXiv171101768O,DBLP:journals/corr/abs-1802-05351,DBLP:journals/corr/abs-1812-11720,yan2018cache,hu2019neural,hong2018security}), we report our main results using the same architecture for both the secret and substitute models.

Further, the adversary has no knowledge of the secret dataset $\mathcal{D}$ on which the model was originally trained. It can however make use of unlabeled public data, i.e.\ the thief dataset $X_\text{thief}$. Note that this data needs to be labeled first by the secret model before it can be used to train the substitute model.

\paragraph{Adversary's goal.}
The goal of the adversary is to obtain a substitute model function that closely resembles (i.e.\ approximates)  the secret model function:
\begin{equation*}
\tilde f \approx f
\end{equation*}
To do so, it trains a substitute model $\tilde f$ on a subset of the thief dataset, $S \subseteq X_\text{thief}$,
\begin{equation*}
\tilde f \approx \arg\min_{f'} \mathcal{L}\big(f', \{(x, f(x)) : x \in S\}\big)
\end{equation*}
where $\mathcal{L}$ is the chosen loss function. As there is a cost associated with querying $f$ and $\left| X_\text{thief} \right|$, the adversary would want $\left| S \right| \ll \left| X_\text{thief} \right|$. The resulting model $\tilde f$ is treated as the extracted model at the end of the process. As it is not possible to arrive at analytical optimum in the general case, the quality of the extracted model is judged using the following \textit{Agreement} metric.

\textbf{Definition (Agreement):}
Two models $f$ and $\tilde f$ agree on the label for a sample $x$ if they predict the same label for the same sample, i.e.\ $f(x) = \tilde f(x)$. The agreement of two networks $f$ and $\tilde f$ is the fraction of samples $x$ from a dataset $\mathcal{D}$ on which they agree, i.e.\ for which $f(x) = \tilde f(x)$
\begin{equation*}
\text{Agreement}(f, \tilde f, \mathcal{D}) = \frac{1}{\left| \mathcal{D} \right|} \sum_{(x, y) \in \mathcal{D}} \mathbbm{1}[f(x) = \tilde f(x)]
\end{equation*}
where $\mathbbm{1}(\cdot)$ is the indicator function. Note that the agreement score does not depend on the true label $y$. Agreement is penalized for every sample for which the predicted labels by the two models $f(x)$ and $\tilde f(x)$ do not match. The higher the agreement between two models on a held-out test set, the more likely it is that the extracted model approximates the secret model well.

\section{Technical details}
\label{sec:technical-details}

\begin{figure*}[t!]
\centering
\scalebox{.95}{
\begin{tikzpicture}[
    >=stealth,
    node distance=3cm,
    block/.style={
      rectangle,
      aspect=0.25,
      draw,
      align=center
    },
  ]
  
  \node[database,label={[align=center]below:\footnotesize Thief dataset\\\footnotesize $X_\text{thief}$},database radius=0.5cm,database segment height=0.25cm] (noise_dataset) at (-1,0) {};
  
  \node[block,minimum width=2cm,minimum height=1.1cm] (seed_noise) at (2.5,0) {\footnotesize Seed samples\\\footnotesize $S_0$};
  %\node[dave,good, minimum size=0.65cm] (original_model) at (6,0) {\footnotesize Secret model $f$};
  \node[block,minimum width=2cm,minimum height=1.1cm] (goodlabeled_noise) at (9.8,0) {\footnotesize True labeled\\ \footnotesize samples $D_{i}$};
  %\node[devil,evil, minimum size=0.65cm] (copied_model) at (13,0) {\footnotesize Substitute model $\tilde f$};

  \node[block,minimum width=1.5cm,minimum height=1.1cm,fill=green!30] (original_model) at (6,0) {\huge \faGears};
  \node at (6, -0.78) {\footnotesize Secret Model $f$};
  \node[block,minimum width=1.5cm,minimum height=1.1cm,fill=orange!30] (copied_model) at (13,0) {\huge \faGears};
  \node at (13, -0.78) {\footnotesize Substitute Model $\tilde f$};

  \node[block,minimum width=2cm,minimum height=1.1cm] (labeled_noise) at (13,-2.3) {\footnotesize Approx.\ labeled\\\footnotesize samples $\tilde D_{i+1}$};
  \node[block,minimum width=2cm,minimum height=1.1cm] (noise_subset) at (6,-2.3) {\footnotesize Next query\\\footnotesize set $S_{i+1}$};
  
  \coordinate (above_noise) at (-1,1.2) {};
  \coordinate (above_copied) at (13,1.2) {};
  
  \node (below_original) at (6,-0.9) {};
  \node (below_copied) at (13,-0.9) {};
  
  \path[->,draw] (noise_dataset) edge node[above,align=center] {\footnotesize \circled{1} \\ \footnotesize Random} (seed_noise);
  \path[] (noise_dataset) edge node[below] {\footnotesize selection} (seed_noise);
  \path[->,draw] (seed_noise) edge node[above] {\footnotesize $s$} (original_model);
  \path[] (seed_noise) edge node[below] {\footnotesize Query} (original_model);
  \path[->,draw] (original_model) edge node[above,align=center] {\footnotesize \circled{2} \\ \footnotesize $(s, f(s))$} (goodlabeled_noise);
  \path[] (original_model) edge node[below] {\footnotesize Collect} (goodlabeled_noise);
  
  \path[->,draw] (goodlabeled_noise) edge node[above,align=center] {\footnotesize \circled{3} \\ \footnotesize Train} (copied_model);
  
  \draw[->] (noise_dataset) -- (above_noise) -- (above_copied) -- (copied_model);
  \path[] (above_copied) edge node[left] {$x$\;} (copied_model);
  \path[] (above_copied) edge node[right] {\footnotesize Query} (copied_model);
  %\path[] (above_noise) edge node[below] {\footnotesize -- All samples $\rightarrow$} (above_copied);
  
  \path[->,draw] (below_copied) edge node[left] {\footnotesize $(x, \tilde f(x))$} (labeled_noise);
  \path[] (below_copied) edge node[right] {\footnotesize Collect \footnotesize \circled{4}} (labeled_noise);
  
  \path[->,draw] (labeled_noise) edge node[above,align=center] {\footnotesize \circled{5} \\ \footnotesize Subset selection strategy} (noise_subset);
  \path[] (labeled_noise) edge node[below] {\footnotesize (e.g., \textsc{k}-center, adversarial, etc.)} (noise_subset);
  
  \path[->,draw] (noise_subset) edge node[left] {\footnotesize $s$\;} (below_original);
  \path[] (noise_subset) edge node[right] {\footnotesize Query} (below_original);
  
\end{tikzpicture}
}
\caption{Our framework for model extraction (see Section~\ref{sec:technical-details} for explanation of steps 1-5).}
\label{fig:framework}
\end{figure*}
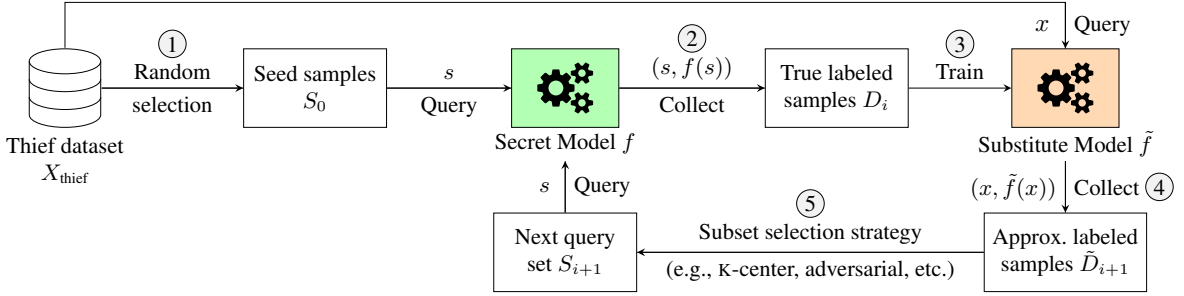

We start with a high-level description of the framework with reference to Figure~\ref{fig:framework}.

\begin{enumerate}
\item The adversary first picks a \todoy{random subset $S_0$ of the unlabeled thief dataset $X_\text{thief}$} to kickstart the process.
\item In the $i$\textsuperscript{th} iteration ($i = 0, 1, 2, \dots, N$), the adversary \todov{queries the samples in $S_i$} against the secret model $f$ and obtains the correctly labeled subset $D_i = \{(x, f(x)) : x \in S_i\}$.
\item Using $D_i$, it trains the substitute model $\tilde f$.
\item The trained substitute model is then queried with all \todoa{samples} in $X_\text{thief}$ to form the approximately labeled dataset $\tilde D_{i+1}$.
\item A subset selection strategy uses $\tilde D_{i+1}$ to select the points $S_{i+1}$ to be queried next.
\end{enumerate}

%\subsection{Extraction algorithm}
\begin{algorithm}[t!]
\caption{Model extraction by active learning}
\label{alg:algorithm}
\SetKwInOut{Input}{Input}
\SetKwInOut{Parameters}{Parameters}
\SetKwInOut{Output}{Output}

\Input{secret model $f$; unlabeled thief dataset $X_\text{thief}$}
\Parameters{\todoa{iteration count} $N$; total query budget $B$;\\\todoy{seed size $k_0$}; validation fraction $\eta$}
\Output{Substitute model, $\tilde f$}

 $\mathrlap{S_\text{valid}}\hphantom{D^X_\text{valid}} \gets \eta B \text{ random datapoints from } X^\text{valid}_\text{thief}$\;
 $\mathrlap{D_\text{valid}}\hphantom{D^X_\text{valid}} \gets \{(x, f(x)) : x \in S_\text{valid}\}$\;
 $\mathrlap{S_0}\hphantom{D^X_\text{valid}} \gets \todoy{k_0} \text{ random datapoints from } X_\text{thief}^\text{train}$\;
 $\mathrlap{D_0}\hphantom{D^X_\text{valid}} \gets \{(x, f(x)) : x \in S_0\}$\;
 $\mathrlap{k}\hphantom{D^X_\text{valid}} \gets \big( (1 - \eta)B - \todoy{k_0} \big) \div n$\;
\For{$i \in \{1 \dots \todoy{N}\}$}{
 $\mathrlap{\tilde f}\hphantom{\tilde D_i} \gets \textsc{TrainNetwork}(D_{i-1}, D_\text{valid})$\;
 $\mathrlap{\tilde D_i}\hphantom{\tilde D_i} \gets \{(x, \tilde f(x)) : x \in X_\text{thief}^\text{train} \land (x,\cdot) \not\in D_{i-1}\}$\;
 $\mathrlap{S_i}\hphantom{\tilde D_i} \gets \textsc{SubsetSelection}(\tilde D_i, D_{i-1}, k)$\;
 $\mathrlap{D_i}\hphantom{\tilde D_i} \gets D_{i-1} \cup \{(x, f(x)) : x \in S_i\}$\;
 }
 \todoy{$\tilde f \gets \textsc{TrainNetwork}(D_N, D_\text{valid})$}\;
\end{algorithm}

The process is repeated for a fixed number of iterations, with the substitute model $\tilde f$ being refined in each iteration. The procedure is formally described in Algorithm~\ref{alg:algorithm}. \todoy{The training procedure followed by \textsc{TrainNetwork} is described in Section~\ref{sec:training}. The details of \textsc{SubsetSelection} follow.}

\subsection{Active learning subset selection strategies}
\label{sec:strategies}
In each iteration, the adversary selects a new set of \todoy{$k$} thief dataset samples $S_i \subseteq X
_\text{thief}$ to label \todov{by querying} the secret model $f$. \todos{This is done using a strategy from the active learning literature:}

%The subset selection strategies used are based both on existing active learning strategies (\texttt{simple}) and their combinations (\texttt{ensemble}). We describe the strategies that we use below:

\begin{itemize}
\item \textit{Random strategy}: A subset of size $k$ consisting of samples $x_n$ is selected uniformly at random, corresponding to pairs $(x_n, \tilde y_n)$ in $\tilde D_{i}$.

\item \textit{Uncertainty strategy}: This method is \todos{based on uncertainty sampling \cite{DBLP:journals/corr/LewisG94}}. For every pair $(x_n, \tilde y_n) \in \tilde{D}_{i}$, the entropy $\mathcal{H}_n$ of predicted probability vectors $\tilde y_n = \tilde f(x_n)$ is computed:
\begin{align*}\mathcal{H}_n = - \sum_{j} \tilde y_{n,j} \log \tilde y_{n,j} \end{align*}where $j$ is the label index. The $k$ samples $x_n$ corresponding to the highest entropy values $\mathcal{H}_n$ (i.e.\ those that the model is least certain about) are selected, breaking ties arbitrarily.
\end{itemize}
\todos{\shortciteauthor{DBLP:journals/corr/abs-1802-09841} demonstrate that the uncertainty strategy does not work well on DNNs. Thus, we also consider two state-of-the-art active learning strategies for DNNs:}

\begin{itemize}
\item \textit{\textsc{k}-center strategy}: We use the greedy \textsc{k}-center algorithm of \todos{\shortciteauthor{sener2017active}} to construct a core-set of samples. This strategy operates in the space of probability vectors produced by the substitute model. The predicted probability vectors \todoy{$\tilde y_m = \tilde f(x_m)$} for samples $\todoy{(x_m, y_m)} \in D_{i-1}$ are considered to be cluster centers. In each iteration, the strategy selects $k$ centers by picking, one at a time, pairs $(x_n, \tilde y_n) \in \tilde D_i$ such that $\tilde y_n$ is the most distant from all existing centers:
\begin{align*}
(x^*_0, \tilde y^*_0) &= \arg \max_{(x_n, \tilde y_n) \in \tilde D_i} \min_{(x_m, y_m) \in D_{i-1}} \todoy{\lVert \tilde y_n - \tilde y_m\rVert_2^2}\\
(x^*_1, \tilde y^*_1) &= \arg \max_{\substack{(x_n, \tilde y_n) \in \tilde D_{i}^1}} \min_{\substack{(x_m, y_m) \in D^1_{i-1}}} \todoy{\lVert \tilde y_n - \tilde y_m\rVert_2^2}
\end{align*}
where:
\begin{align*}
\tilde D_i^1 \gets& \tilde D_i \setminus \{(x^*_0, \tilde y^*_0)\}\\
D_{i-1}^1 \gets& D_{i-1} \cup \{\big(x^*_0, f(x^*_0)\big)\}
\end{align*}
i.e.\ $(x^*_0, \tilde y^*_0)$ is moved to the set of selected centers. This process is repeated to obtain $k$ pairs. The samples $x^*_0, x^*_1, \dots x^*_k$ corresponding to the chosen pairs are selected.

\item \textit{Adversarial strategy}: We use the DeepFool Active Learning (DFAL) algorithm by \todos{\shortciteauthor{DBLP:journals/corr/abs-1802-09841}}. In this strategy, DeepFool \cite{moosavi2016deepfool} (explained in Section~\ref{sec:deepfool}) is applied to every sample $x_n \in \tilde D_i$ to obtain a perturbed $\hat x_n$ that gets misclassified by the substitute model $\tilde f$, i.e.\ $\tilde f(x_n) \neq \tilde f(\hat x_n)$. (Note that this does not involve querying the secret model.) Let: $$\alpha_n = \lVert x_n - \hat x_n \rVert^2_2$$

% DeepFool \cite{ducoffe2018adversarial}

%%%% MOVED THIS TABLE HERE TO FORCE SPACING 
\begin{table*}[t!]
	\caption{Details of datasets for image and text classification tasks. \# Train, \# Val and \# Test refer to the number of samples in the train, validation and test folds respectively. Note that the thief datasets (ImageNet subset and WikiText-2) do not have predefined folds, but the fractions used for training and validation have been tabulated for reference.}
	\label{tab:datasets}
	\begin{subtable}[t]{\linewidth}
		\centering
		\caption{Details of datasets for image classification tasks.}
		\label{tab:datasets-image}
		\begin{tabular}{lrrrrr}  
			\toprule
			\todos{Image} Dataset & Dimensions & \# Train & \# Val & \# Test & \# Classes \\
			\midrule
			MNIST         & $28 \times 28 \times 1$ & 48K & 12K & 10K & 10 \\
			F-MNIST & $28 \times 28 \times 1$ & 48K & 12K & 10K & 10 \\
			CIFAR-10      & $32 \times 32 \times 3$ & 40K & 10K & 10K & 10 \\
			GTSRB         & \todoy{$32 \times 32 \times 3$} & $\sim$ 31K & $\sim$ 8K & $\sim$ 12K & 43 \\
			\midrule
			ImageNet subset & $64 \times 64 \times 3$ & 100K & 50K & -- & -- \\
			\bottomrule
		\end{tabular}
	\end{subtable}\bigskip\newline
	\begin{subtable}[t]{\linewidth}
		\centering
		\caption{Details of datasets for text classification tasks.}
		\label{tab:datasets-text}
		\begin{tabular}{lrrrr}  
			\toprule
			\todos{Text} Dataset & \# Train & \# Val & \# Test & \# Classes \\
			\midrule
			MR & 7,676 & 1,920 & 1,066 & 2 \\
			IMDB         & 20K & 5K & 25K & 2 \\
			AG News    & 96K & 24K & $\sim$ 7K & 5 \\
			QC           & $\sim$ 12K & 3K & .5K & 6 \\
			\midrule
			WikiText-2 & $\sim$ 89K & $\sim$ 10K & -- & -- \\
			\bottomrule
		\end{tabular}
	\end{subtable}
	
\end{table*}

DFAL is a margin-based approach to active learning, i.e.\ it identifies samples that lie close to the decision boundary. To do this, it prefers samples $x_n$ corresponding to lower values of $\alpha_n$, i.e.\ smallest distance between $x_n$ and its adversarially perturbed neighbor $\hat x_n$ that lies across the decision boundary. Thus, this strategy selects the $k$ samples $x_n$ corresponding to the lowest perturbation $\alpha_n$.
\end{itemize}

\subsection{Ensemble of subset selection strategies}

\todos{While the \textsc{k}-center strategy maximizes diversity, it does not ensure that each individual sample is helpful to the learner. On the contrary, while the adversarial strategy ensures that each individual sample is informative, it does nothing to eliminate redundancy across the samples selected. Inspired by this observation, we introduce the following \textit{ensemble} subset selection strategy called \textit{Adversarial+\textsc{k}-center strategy}.}

In this ensemble strategy, the adversarial strategy is first used to \todov{pick $\rho$ points} ($\rho$ is a configurable parameter). Of these, $k$ points are selected using the \textsc{k}-center strategy.\todos{The adversarial strategy first picks samples that lie close to the decision boundary. Following this, the \textsc{k}-center strategy selects a subset of these points with an aim to maximize diversity. We demonstrate the effectiveness of this strategy experimentally in Section~\ref{sec:image-results}.}

\begin{figure*}[t!]
	\centering
	\begin{tikzpicture}[
	>=stealth,
	node distance=3cm,
	block/.style={
		rectangle,
		aspect=0.25,
		draw,
		align=center
	},
	]
	
	\node[block,minimum width=50,rotate=90] (input) at (-0.5,0) {\footnotesize Input};
	
	\node[block,minimum width=50,rotate=90,fill=red!20] (conv11a) at (1,0) {\footnotesize conv1.1a};
	\node[block,minimum width=50,rotate=90,fill=red!20] (conv11b) at (1.5,0) {\footnotesize conv1.1b};
	\node[block,minimum width=50,rotate=90,fill=red!5] (pool11) at (2.03,0) {\footnotesize pool1.1};
	
	\node[block,minimum width=50,rotate=90,fill=red!20] (conv12a) at (3,0) {\footnotesize conv1.2a};
	\node[block,minimum width=50,rotate=90,fill=red!20] (conv12b) at (3.5,0) {\footnotesize conv1.2b};
	\node[block,minimum width=50,rotate=90,fill=red!5] (pool12) at (4.03,0) {\footnotesize pool1.2};
	
	\node[block,minimum width=50,rotate=90,fill=green!20] (conv21a) at (5,0) {\footnotesize conv2.1a};
	\node[block,minimum width=50,rotate=90,fill=green!20] (conv21b) at (5.5,0) {\footnotesize conv2.1b};
	\node[block,minimum width=50,rotate=90,fill=green!5] (pool21) at (6.03,0) {\footnotesize pool2.1};
	
	\node[block,minimum width=50,rotate=90,fill=green!20] (conv22a) at (7,0) {\footnotesize conv2.2a};
	\node[block,minimum width=50,rotate=90,fill=green!20] (conv22b) at (7.5,0) {\footnotesize conv2.2b};
	\node[block,minimum width=50,rotate=90,fill=green!5] (pool22) at (8.03,0) {\footnotesize pool2.2};
	
	\node[block,minimum width=50,rotate=90,fill=blue!20] (conv31a) at (9,0) {\footnotesize conv3.1a};
	\node[block,minimum width=50,rotate=90,fill=blue!20] (conv31b) at (9.5,0) {\footnotesize conv3.1b};
	\node[block,minimum width=50,rotate=90,fill=blue!5] (pool31) at (10.03,0) {\footnotesize pool3.1};
	
	\node[block,minimum width=50,rotate=90,fill=blue!20] (conv32a) at (11,0) {\footnotesize conv3.2a};
	\node[block,minimum width=50,rotate=90,fill=blue!20] (conv32b) at (11.5,0) {\footnotesize conv3.2b};
	\node[block,minimum width=50,rotate=90,fill=blue!5] (pool32) at (12.03,0) {\footnotesize pool3.2};
	
	\node[block,minimum width=50,rotate=90,fill=gray!20] (fc) at (13,0) {\footnotesize FC};
	\node[block,minimum width=50,rotate=90] (end) at (15,0) {\footnotesize Prob};
	
	\node at (1.5,-1.2) {\begin{tabular}{c}
		\scriptsize $\mathbf{32}$ filters each
		\end{tabular}};
	
	\node at (3.5,-1.2) {\begin{tabular}{c}
		\scriptsize $\mathbf{32}$ filters each
		\end{tabular}};
	
	\node at (5.5,-1.2) {\begin{tabular}{c}
		\scriptsize $\mathbf{64}$ filters each
		\end{tabular}};
	
	\node at (7.5,-1.2) {\begin{tabular}{c}
		\scriptsize $\mathbf{64}$ filters each
		\end{tabular}};
	
	\node at (9.5,-1.2) {\begin{tabular}{c} 
		\scriptsize $\mathbf{128}$ filters each
		\end{tabular}};
	
	\node at (11.5,-1.2) {\begin{tabular}{c} 
		\scriptsize $\mathbf{128}$ filters each
		\end{tabular}};
	
	\node at (13,-1.2) {\begin{tabular}{c} 
		\scriptsize Projection
		\end{tabular}};
	
	\draw[->] (input) -- (conv11a);
	\draw[-] (conv11a) -- (conv11b);
	\draw[-] (conv11b) -- (pool11);
	
	\draw[->] (pool11) -- (conv12a);
	\draw[-] (conv12a) -- (conv12b);
	\draw[-] (conv12b) -- (pool12);
	
	\path[->,draw] (pool12) edge node[above] {} (conv21a);  
	\draw[-] (conv21a) -- (conv21b);
	\draw[-] (conv21b) -- (pool21);
	
	\draw[->] (pool21) -- (conv22a);
	\draw[-] (conv22a) -- (conv22b);
	\draw[-] (conv22b) -- (pool22);
	
	\path[->,draw] (pool22) edge node[above] {} (conv31a);  
	\draw[-] (conv31a) -- (conv31b);
	\draw[-] (conv31b) -- (pool31);
	
	\draw[->] (pool31) -- (conv32a);
	\draw[-] (conv32a) -- (conv32b);
	\draw[-] (conv32b) -- (pool32);
	
	\path[->,draw] (pool32) edge node[above] {} (fc);  
	\path[->,draw] (fc) edge node[above] {\footnotesize Softmax} (end);
	\end{tikzpicture}
	\caption{Network architecture for image classification tasks}.
	\label{fig:network}
\end{figure*}
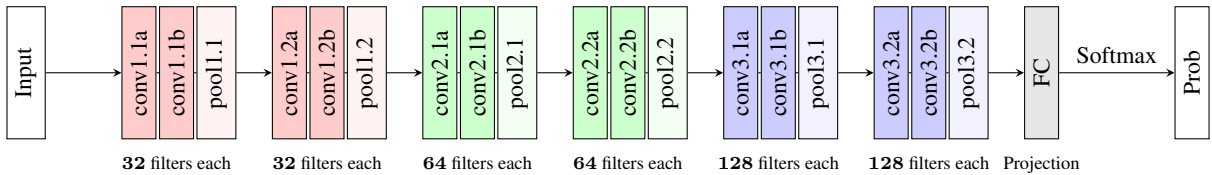

\section{Experimental setup}
We now describe the datasets and DNN architecture used in our experiments.

\subsection{Datasets}
\label{sec:datasets}

The details of each dataset can be found in Table~\ref{tab:datasets}.

\paragraph{Secret datasets\todov{.}}
For image classification, we use the MNIST dataset of handwritten digits \cite{lecun1998gradient}, the Fashion-MNIST (F-MNIST) dataset of small grayscale images of fashion products across 10 categories \cite{DBLP:journals/corr/abs-1708-07747}, the CIFAR-10 dataset of tiny color images \cite{krizhevsky2009learning} and the German Traffic Sign Recognition Benchmark (GTSRB) \cite{stallkamp2012man}.

For text classification, we use the MR dataset \cite{Pang+Lee:05a} of 5,331 positive and 5,331 statements from movie reviews, the IMDB \cite{maas-EtAl:2011:ACL-HLT2011} dataset of movie reviews, AG News corpus \footnote{\url{https://di.unipi.it/~gulli/AG_corpus_of_news_articles.html}} of news from 5 categories and the QC question classification dataset \cite{li2002learning}.

\paragraph{Thief dataset\todov{.}}
For images, we use a subset of the ILSVRC2012-14 dataset \cite{russakovsky2015ImageNet} as the thief dataset. In particular, we use a downsampled version of this data prepared by \shortciteauthor{DBLP:journals/corr/ChrabaszczLH17}. The training and validation splits are reduced to a subset of size 100,000, while the test split is left unchanged. % The samples are rescaled or made greyscale if required before they are fed into the network.

For text, we use sentences extracted from the WikiText-2 \cite{DBLP:journals/corr/MerityXBS16} dataset of Wikipedia articles.

\subsection{DNN architecture}
\label{sec:architecture}
The same \textbf{base complexity} architectures are used for both the secret and the substitute model for our primary evaluation in Sections~\ref{sec:image-results} and \ref{sec:text-results}. We also conduct additional experiments on image classification tasks where the model complexities are varied between the secret and substitute models in Section~\ref{sec:grids}.

We first describe the base complexity architectures for image and text classification:

\paragraph{Image classification\todov{.}} We use a multi-layered CNN, shown in Figure~\ref{fig:network}. The input is followed by 3 convolution blocks. Each convolution block consists of 2 repeated units -- a single repeated unit consists of 2 convolution ($3\times 3$ kernel with stride $1$) and 1 pooling ($2\times 2$ kernel with stride $2$) layers. Each convolution is followed by a ReLU activation \todoy{and batch normalization layer}. Pooling is followed by a dropout. \todoy{Convolution layers in each block use 32, 64 and 128 filters respectively.} No two layers share parameters. The output of the final pooling layer is flattened and passed through fully connected and softmax layers to obtain the vector of output probabilities.

% batch norm citation \cite{ioffe2015batch}

\paragraph{Text classification\todov{.}} We use the CNN for sentence classification by \todos{\shortciteauthor{kim2014convolutional}}. In the secret model, word2vec \cite{mikolov2013efficient} is first used to obtain the word embeddings. The embeddings are then concatenated and 100 1-dimensional filters each of sizes 3, 4 and 5 are applied to convolve over time. This is followed by max-over-time pooling, which produces a 300-dimensional vector. This vector is then passed through fully connected and softmax layers to obtain the vector of output probabilities.

 % There is work on discovering the underlying architecture and hyperparameters of a model with only black-box access to it \cite{DBLP:journals/corr/abs-1802-05351,2017arXiv171101768O}.

\subsection{Training Regime}
\label{sec:training}

For training, we use the Adam optimizer \cite{DBLP:journals/corr/KingmaB14} with default hyperparameters ($\beta_1 = 0.9$, $\beta_2 = 0.999$, $\epsilon = 10^{-8}$ and a learning rate of $0.001$). In each iteration, the network is trained starting from the same random initialization for at most 1,000 epochs with a batch size of \todoa{150 (for images) or 50 (for text)}. Early stopping is used with a patience of \todoa{100 epochs (for images) or 20 epochs (for text)}. An $L_2$ regularizer is applied to all the model parameters with a loss term multiplier of $0.001$, and dropout is applied at a rate of 0.1 for all datasets other than CIFAR-10. For CIFAR-10, a dropout of 0.2 is used. At the end of each epoch, the model is evaluated and the $F_1$ measure on the validation split is recorded. The model with the best validation $F_1$ measure is \todov{selected as $\tilde f$ in that iteration}.

Our experiments are run on a server with a 24-core Intel Xeon Gold 6150 CPU and NVIDIA GeForce GTX 1080Ti GPUs. We use the algorithm parameters \todoy{$k_0 = 0.1 B$} (where $B$ is the total query budget, as in Algorithm~\ref{alg:algorithm}) and $\eta = 0.2$ across all our experiments. \todov{For the ensemble strategy, we set $\rho = B$, the total query budget.}

\section{Experimental results}
In our experiments we seek to obtain answers to the following questions:
\begin{enumerate}
\item
	How do the various active learning algorithms compare in their performance, i.e., in terms of the agreement between the secret model and substitute model?
\item 
	How does the query budget affect agreement?
\item 
	What is the impact of using universal thief datasets over using uniform noise samples to query the secret model?
\item 
	What is the impact of the DNN architectures (of the secret and substitute models) on the agreement obtained?
	
\end{enumerate}

The first three questions are answered in the context of image datasets in Section~\ref{sec:image-results} and text datasets in Section~\ref{sec:text-results}. The fourth question is answered in Section~\ref{sec:grids}.

In our experiments, for all but the \textit{random} strategy, training is done iteratively. As the choice of samples in random strategy is not affected by the substitute model \todoy{$\tilde f$} obtained in each iteration, we skip iterative training. We also train a substitute model using the full thief dataset for comparison. The metric used for evaluation of the closeness between the secret model $f$ and the substitute model $\tilde f$ is agreement between $f$ and $\tilde f$, evaluated on the test split of the secret dataset.

%\theoremstyle{definition}
%\begin{definition}{\textit{Agreement}}

%\end{definition}

%As a metric, we prefer agreement to accuracy as it is a better indicator that. If a model is truly extracted, $\tilde f$ should make the same mistakes as $f$, not just learn to do well on the task it was originally trained to perform.

\subsection{Image classification}
\label{sec:image-results}

\begin{figure*}[t!]
\begin{subfigure}[t]{0.5\linewidth}
\begin{tikzpicture}[scale = 0.95]
\begin{axis}[legend pos=south east,width=9.5cm,height=4.5cm,xmin=-1,xmax=11]
\plotfile{mnistdata.dat}
\legend{Uncertainty,\textsc{k}-center,Adversarial,Adv+\textsc{k}-cen}
\addplot[mark=none, black, dashed,domain=-1:11] {.9590};
\addlegendentry{Random}
\end{axis}
\end{tikzpicture}
\caption{MNIST dataset}
\label{tab:iterations-mnist}
\end{subfigure}%
\begin{subfigure}[t]{0.5\linewidth}
\begin{tikzpicture}[scale = 0.95]
\begin{axis}[legend pos=south east,width=9.5cm,height=4.5cm,xmin=-1,xmax=11]
\plotfile{fashionmnistdata.dat}
\legend{Uncertainty,\textsc{k}-center,Adversarial,Adv+\textsc{k}-cen}
\addplot[mark=none, black, dashed,domain=-1:11] {.6932};
\addlegendentry{Random}
\legend{}
\end{axis}
\end{tikzpicture}
\caption{F-MNIST dataset}
\label{tab:iterations-fashionmnist}
\end{subfigure}\par\bigskip
\begin{subfigure}[t]{0.5\linewidth}
\begin{tikzpicture}[scale = 0.95]
\begin{axis}[legend pos=south east,width=9.5cm,height=4.5cm,xmin=-1,xmax=11]
\plotfile{cifar10data.dat}
\legend{Uncertainty,\textsc{k}-center,Adversarial,Adv+\textsc{k}-cen}
\addplot[mark=none, black, dashed,domain=-1:11] {.6932};
\addlegendentry{Random}
\legend{}
\end{axis}
\end{tikzpicture}
\caption{CIFAR-10 dataset}
\label{tab:iterations-fashionmnist}
\end{subfigure}%
\begin{subfigure}[t]{0.5\linewidth}
\begin{tikzpicture}[scale = 0.95]
\begin{axis}[legend pos=south east,width=9.5cm,height=4.5cm,xmin=-1,xmax=11]
\plotfile{gtsrdata.dat}
\legend{Uncertainty,\textsc{k}-center,Adversarial,Adv+\textsc{k}-cen}
\addplot[mark=none, black, dashed,domain=-1:11] {.6932};
\addlegendentry{Random}
\legend{}
\end{axis}
\end{tikzpicture}
\caption{GTSRB dataset}
\label{tab:iterations-fashionmnist}
\end{subfigure}
\caption{The improvement in agreement for image classification experiments with a total budget of 20K over 10 iterations. Since random is not run iteratively, it is indicated as a line parallel to the X-axis.}
\label{fig:iterations}
\end{figure*}
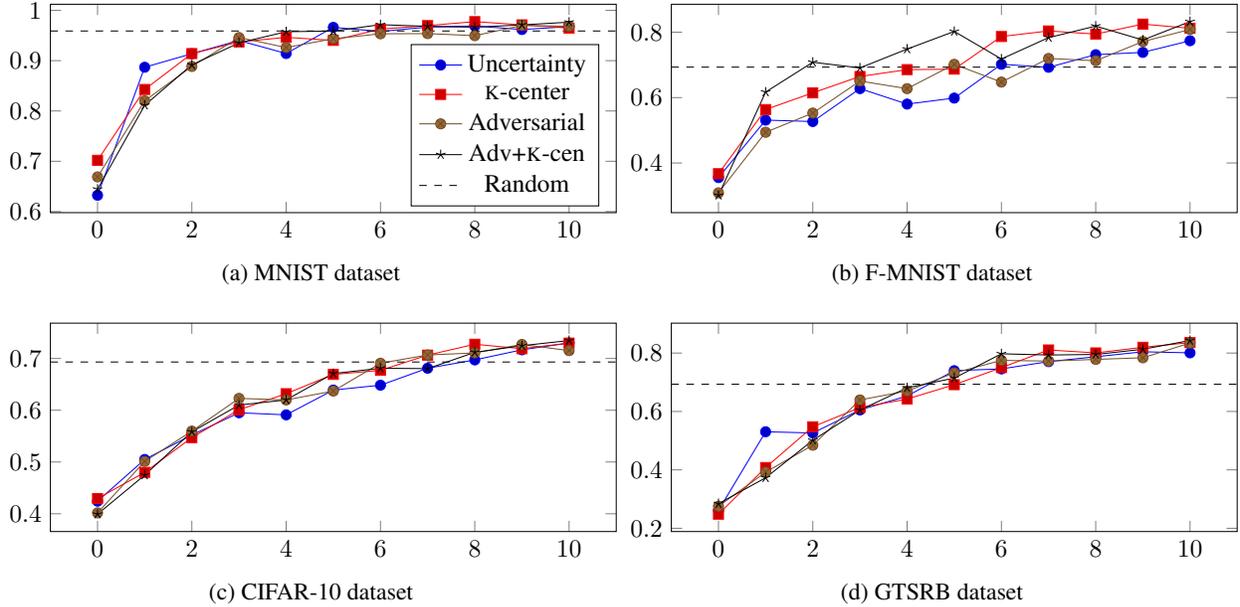

\begin{table*}[t!]
\caption{The agreement on the secret test set for image classification tasks. Each row corresponds to a subset selection strategy, while each column corresponds to a query budget.}
\label{tab:results}
\begin{subfigure}[t]{0.5\linewidth}
\centering
\caption{MNIST dataset}
\label{tab:results-mnist}
\begin{tabular}{lrrrrr}
	\toprule
	%& \multicolumn{5}{c}{Substitute model agreement (\%)}\\
	Strategy & 10K & 15K & 20K & 25K & 30K\\
	\midrule
	Random & 91.64 & 95.19 & 95.90 & 97.48 & 97.36\\
	Uncertainty & 94.64 & 97.43 & 96.77 & 97.29 & 97.38\\
	\textsc{k}-center & \textbf{95.80} & 95.66 & 96.47 & \textbf{97.81} & 97.95\\
	Adversarial & 95.75 & 95.59 & 96.84 & 97.74 & 97.80\\
	Adv+\textsc{k}-cen & 95.40 & \textbf{97.64} & \textbf{97.65} & 97.60 & \textbf{98.18}\\
	\midrule
	\multicolumn{5}{l}{Using the full thief dataset (100K):}  & 98.81\\
	\multicolumn{5}{l}{Using uniform noise samples (100K):}  & 20.56\\
	\bottomrule
\end{tabular}
\end{subfigure}%
\begin{subfigure}[t]{0.5\linewidth}
\centering
\caption{F-MNIST dataset}
\label{tab:results-fashionmnist}
\begin{tabular}{lrrrrr}
	\toprule
Strategy & 10K & 15K & 20K & 25K & 30K\\
\midrule
Random & 62.36 & 67.61 & 69.32 & 71.76 & 71.57\\
Uncertainty & 71.18 & 72.19 & 77.39 & 77.88 & 82.63\\
\textsc{k}-center & 71.37 & 77.03 & 81.21 & 79.46 & 82.90\\
Adversarial & 67.61 & 69.89 & 80.84 & 80.28 & 81.17\\
Adv+\textsc{k}-cen & \textbf{73.51} & \textbf{81.45} & \textbf{83.24} & \textbf{80.83} & \textbf{83.38}\\
\midrule
\multicolumn{5}{l}{Using the full thief dataset (100K):}  & 84.17\\
\multicolumn{5}{l}{Using uniform noise samples (100K):}  & 17.55\\
	\bottomrule
\end{tabular}
\end{subfigure}\par\bigskip
\begin{subfigure}[t]{0.5\linewidth}
\centering
\caption{CIFAR-10 dataset}
\label{tab:results-cifar}
\begin{tabular}{lrrrrr}
	\toprule
	Strategy & 10K & 15K & 20K & 25K & 30K\\
	\midrule
	Random & 63.75 & 68.93 & 71.38 & 75.33 & 76.82\\
	Uncertainty & 63.36 & 69.45 & 72.99 & 74.22 & 76.75\\
	\textsc{k}-center & \textbf{64.20} & 70.95 & 72.97 & 74.71 & 78.26\\
	Adversarial & 62.49 & 68.37 & 71.52 & \textbf{77.41} & 77.00\\
	Adv+\textsc{k}-cen & 61.52 & \textbf{71.14} & \textbf{73.47} & 74.23 & \textbf{78.36}\\
	\midrule
	\multicolumn{5}{l}{Using the full thief dataset (100K):}  & 81.57\\
	\multicolumn{5}{l}{Using uniform noise samples (100K):}  & 10.62\\
	\bottomrule
\end{tabular}
\end{subfigure}%
\begin{subfigure}[t]{0.5\linewidth}
\centering
\caption{GTSRB dataset}
\label{tab:results-gtsr}
\begin{tabular}{lrrrrr}
	\toprule
	Strategy & 10K & 15K & 20K & 25K & 30K\\
	\midrule
	Random & 67.72 & 77.71 & 79.49 & 82.14 & 83.84\\
	Uncertainty & 67.30 & 73.92 & 80.07 & 83.61 & 85.49\\
	\textsc{k}-center & 70.89 & \textbf{81.03} & 83.59 & \textbf{85.81} & 85.93\\
	Adversarial & \textbf{72.71} & 79.44 & 83.43 & 84.41 & 83.98\\
	Adv+\textsc{k}-cen & 70.79 & 79.55 & \textbf{84.29} & 85.41 & \textbf{86.71}\\
	\midrule
	\multicolumn{5}{l}{Using the full thief dataset (100K):}  & 91.42\\
	\multicolumn{5}{l}{Using uniform noise samples (100K):}  & 45.53\\
	\bottomrule
\end{tabular}
\end{subfigure}
\end{table*}

For each image dataset (described in Section~\ref{sec:datasets}), we run our framework across the following total query budgets: 10K, 15K, 20K, 25K and 30K \todoy{(K = 1,000)}. For \todov{a budget of} 20K, we show the agreement at the end of each iteration for every strategy and each dataset in Figure~\ref{fig:iterations}.

We tabulate the agreement obtained at the end of the final iteration for each experiment in Table~\ref{tab:results}. Our observations across these 20 experiments are as follows:

\paragraph{\clearcolor{Effectiveness of active learning}\todov{.}} \todoy{The benefits of careful selection of thief dataset samples can be clearly seen:} there is no dataset for which the random strategy performs better than all of the other strategies. In particular, \textsc{k}-center underperforms only once, while adversarial and adversarial+\textsc{k}-center underperform twice. \todos{Uncertainty} underperforms 6 times, but this is in line with the findings of \todos{\shortciteauthor{DBLP:journals/corr/abs-1802-09841}}.

\paragraph{Effectiveness of the ensemble method\todov{.}} The agreement of the models is improved by the ensemble strategy over the basic adversarial strategy in 14 experiments. Of these, the ensemble strategy emerges as the winner in 13 experiments -- a clear majority. \todoy{This improvement in agreement bears evidence to the increased potential of the combined strategy in extracting information from the secret model.} The other competitive method is the \textsc{k}-center method, which wins in 5 experiments. This is followed by the adversarial strategy which won in 2 experiments.

\paragraph{Impact of the number of iterations\todov{.}} Table~\ref{tab:num_iter_softmax} shows that with an increase in the number of iterations, there is an improvement in agreement for the same budget. \todoy{Thus, the substitute model agreement can be improved by increasing the number of iterations (at the expense of increased training time).}

\begin{table}[t!]
	\centering
	\caption{\todoa{Agreement on the secret test set for each dataset (total budget of 10K). Here, $N$ refers to the number of iterations (as in Algorithm~\ref{alg:algorithm}). \textbf{Top-1} refers to the adversary having access only to the top prediction, while in \textbf{Softmax}, they have access to the output probability distribution. The agreement is reported for the winning strategy in each case.}}
	\label{tab:num_iter_softmax}
	\bigskip
	
	\begin{tabular}{lrrr}  
		\toprule
		Dataset & \multicolumn{3}{c}{Substitute model agreement (\%)} \\
		\midrule
		& Top-1 & Top-1 & Softmax \\
		& $N = 10$ & $N = 20$ & $N = 10$ \\
		\midrule
		MNIST         & 95.80 & 96.74 & 98.61 \\
		F-MNIST & 73.51 & 78.84 & 82.13  \\
		CIFAR-10      & 64.20 & 64.23 & 77.29  \\
		GTSRB         & 72.71 & 72.78 & 86.90  \\
		\bottomrule
	\end{tabular}
\end{table}

\paragraph{Impact of access to output probability distribution\todov{.}} Table~\ref{tab:num_iter_softmax} demonstrates that access to the output probabilities of the secret model results in an improvement in agreement. \todoy{We believe that this is because the substitute model receives a signal corresponding to every output neuron for each thief dataset sample that it is trained on. Consequently, the substitute model learns a better approximation. However,} as many MLaaS models return only the Top-\textsc{K} or often Top-1 prediction, we run our experiments on the more restricted setting with access to only the Top-1 prediction.

\paragraph{\clearcolor{Impact of the query budget}\todov{.}} \todoy{As is evident from Table \ref{tab:results}, there is almost always a substantial improvement in agreement when increasing the total query budget.}

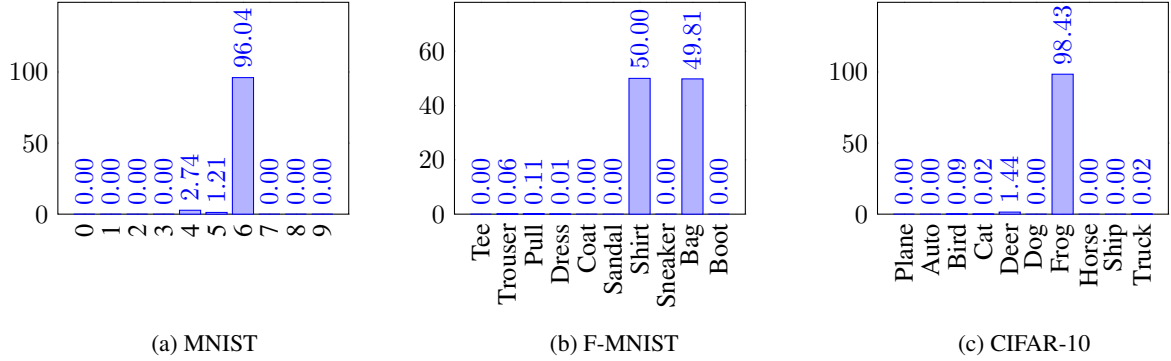
\begin{figure*}[t!]
\begin{subfigure}[t]{0.33\linewidth}

\begin{tikzpicture}
\centering
\begin{axis}[
    ybar,
    xmin=0,
    ymin=0,
    ymax=149,
    xtick=data,
    width=\linewidth,
    height=4.4cm,
    nodes near coords,
    nodes near coords align={vertical},
    every node near coord/.append style={rotate=90, anchor=west},
    every node near coord/.append style={
            /pgf/number format/fixed,
            /pgf/number format/precision=2,
            /pgf/number format/fixed zerofill
    },
    bar width = 8pt,
    major tick length=0cm,
	xtick={1,2,3,4,5,6,7,8,9,10},
	xticklabels={\hphantom{Sneaker}\llap{0},\hphantom{Sneaker}\llap{1},2,3,4,5,6,7,8,9},
    xticklabel style={rotate=90},
    ]
\addplot table [x=a, y=c, col sep=comma] {mnist-data.csv};
\end{axis}
\end{tikzpicture}
\caption{MNIST}
\end{subfigure}%
\begin{subfigure}[t]{0.33\linewidth}

\begin{tikzpicture}
\centering
\begin{axis}[
    ybar,
    xmin=0,
    ymin=0,
    ymax=78,
    xtick=data,
    height=4.4cm,
    width=\linewidth,
    nodes near coords,
    nodes near coords align={vertical},
    every node near coord/.append style={rotate=90, anchor=west},
    every node near coord/.append style={
            /pgf/number format/fixed,
            /pgf/number format/precision=2,
            /pgf/number format/fixed zerofill
},
    bar width = 8pt,
    major tick length=0cm,
	xtick={1,2,3,4,5,6,7,8,9,10},
	xticklabels={\hphantom{Sneaker}\llap{Tee},Trouser,Pull,Dress,Coat,Sandal,Shirt,Sneaker,Bag,Boot},
    xticklabel style={rotate=90},
    ]
\addplot table [x=a, y=c, col sep=comma] {fashionmnist-data.csv};
\end{axis}
\end{tikzpicture}
\caption{F-MNIST}
\end{subfigure}%
\begin{subfigure}[t]{0.33\linewidth}

\begin{tikzpicture}
\centering
\begin{axis}[
    ybar,
    xmin=0,
    ymin=0,
    ymax=149,
    xtick=data,
    height=4.4cm,
    width=\linewidth,
    nodes near coords,
    nodes near coords align={vertical},
    every node near coord/.append style={rotate=90, anchor=west},
    every node near coord/.append style={
            /pgf/number format/fixed,
            /pgf/number format/precision=2,
            /pgf/number format/fixed zerofill
},
    bar width = 8pt,
    major tick length=0cm,
	xtick={1,2,3,4,5,6,7,8,9,10},
	xticklabels={\hphantom{Sneaker}\llap{Plane},Auto,Bird,Cat,Deer,Dog,Frog,Horse,Ship,Truck},
    xticklabel style={rotate=90},
    ]
\addplot table [x=a, y=c, col sep=comma] {cifar-data.csv};
\end{axis}
\end{tikzpicture}
\caption{CIFAR-10}
\end{subfigure}
\caption{The distribution of labels (frequency in \%) assigned by the secret model to uniform noise input.}
\label{fig:uniform}
\end{figure*}

\paragraph{Effectiveness of universal thief datasets\todov{.}} We see that uniform noise (as used in prior work by \shortciteauthor{DBLP:journals/corr/TramerZJRR16}) achieves a low agreement on all datasets. \todoy{The reason for failure is as follows: in our experiments, we observe that when the secret model is queried with uniform noise, there are many labels which are predicted extremely rarely, while others dominate, e.g., the digit \textit{6} dominates in MNIST and \textit{Frog} dominates in CIFAR-10 (see Figure~\ref{fig:uniform}). In other words, it is difficult for an adversary to discover images belonging to certain classes using uniform noise. This problem is alleviated via the use of universal thief datasets like ImageNet}. On an average, using the full thief dataset (100K) leads to an improvement in agreement by $\mathbf{4.82}\times$ over the uniform baseline. Even with a budget of 30K, an improvement of $\mathbf{4.70}\times$ is retained with active learning.
%The use of a line search technique as in \cite{DBLP:journals/corr/TramerZJRR16} does not help.

%%%%%%%%%%%%%%%%%%%%%%%%%%%%%%%%%%%%%%%%%%%%%%%%%
% START OF NOISE RESULTS
%%%%%%%%%%%%%%%%%%%%%%%%%%%%%%%%%%%%%%%%%%%%%%%%%

%\paragraph{Effectiveness of universal thief.} 
% The universal thief alleviates the problem of extreme class imbalance. Even without active learning, the random strategy significantly outperforms the uniform noise baseline.

\subsection{Influence of substitute model architecture}
\label{sec:grids}

To check the influence of the architecture on the substitute model, we consider the following three options:
\begin{itemize}
       \item \textbf{Lower complexity (LC) architecture:} This DNN architecture has two convolution blocks, with two repeated units each (consisting of two convolution layers, followed by a pooling layer). The convolution layers in each block have 32 and 64 filters, respectively.
       \item \textbf{Base complexity (BC) architecture:} This architecture has three convolution blocks, with three repeated units each (of the same configuration). The convolution layers in each block have 32, 64 and 128 filters, respectively. This is the architecture described in Section~\ref{sec:architecture} and used in all the other experiments.
       \item \textbf{Higher complexity (HC) architecture:} This architecture has four convolution blocks, with two repeated units each (of the same configuration). The convolution layers in each block have 32, 64, 128 and 256 filters, respectively.
\end{itemize}

\begin{table*}[t!]
\caption{The agreement on the secret test set for image classification tasks, when architectures of different complexity are used as the secret model and substitute model. Each row corresponds to a secret model architecture, while each column corresponds to a substitute model architecture.}
\label{tab:grids}
\begin{subfigure}[t]{0.5\linewidth}
\centering
\caption{MNIST dataset}
\label{tab:results-mnist}
\begin{tabular}{lrrr}
	\toprule
	%& \multicolumn{5}{c}{Substitute model agreement (\%)}\\
	& \multicolumn{3}{c}{Substitute model}\\
	Secret model & LC & BC & HC\\
	\midrule
	Lower Complexity (LC) & \textbf{98.73} & 98.15 & 97.63 \\
	Base Complexity (BC) & 97.21 & \textbf{98.81} & 98.10 \\
	Higher Complexity (HC) & 96.75 & 98.05 & \textbf{98.36} \\
	\bottomrule
\end{tabular}
\end{subfigure}%
\begin{subfigure}[t]{0.5\linewidth}
\centering
\caption{F-MNIST dataset}
\label{tab:results-fashionmnist}
\begin{tabular}{lrrr}
	\toprule
	%& \multicolumn{5}{c}{Substitute model agreement (\%)}\\
	& \multicolumn{3}{c}{Substitute model}\\
	Secret model & LC & BC & HC\\
	\midrule
	Lower Complexity (LC) & \textbf{87.15} & 80.15 & 75.26 \\
	Base Complexity (BC) & 	81.50 & \textbf{84.17} & 79.88 \\
	Higher Complexity (HC) & 79.83 & 73.35 & \textbf{84.01} \\
	\bottomrule
\end{tabular}
\end{subfigure}\par\bigskip
\begin{subfigure}[t]{0.5\linewidth}
\centering
\caption{CIFAR-10 dataset}
\label{tab:results-cifar}
\begin{tabular}{lrrr}
	\toprule
	%& \multicolumn{5}{c}{Substitute model agreement (\%)}\\
	& \multicolumn{3}{c}{Substitute model}\\
	Secret model & LC & BC & HC\\
	\midrule
	Lower Complexity (LC) & \textbf{78.34} & 76.83 & 74.48 \\
	Base Complexity (BC) & 80.66 & 81.57 & \textbf{81.80} \\
	Higher Complexity (HC) & 74.34 & \textbf{79.17} & 78.82 \\
	\bottomrule
\end{tabular}
\end{subfigure}%
\begin{subfigure}[t]{0.5\linewidth}
\centering
\caption{GTSRB dataset}
\label{tab:results-gtsr}
\begin{tabular}{lrrr}
	\toprule
	%& \multicolumn{5}{c}{Substitute model agreement (\%)}\\
	& \multicolumn{3}{c}{Substitute model}\\
	Secret model & LC & BC & HC\\
	\midrule
	Lower Complexity (LC) & \textbf{95.02} & 92.30 & 86.88 \\
	Base Complexity (BC) & 90.08 & \textbf{91.42} & 91.28 \\
	Higher Complexity (HC) & 80.95 & \textbf{86.50} & 84.69 \\
	\bottomrule
\end{tabular}
\end{subfigure}
\end{table*}

We consider all possible combinations of the above DNN architectures applied to both the secret and substitute models. The results of our experiments on the image classification tasks using all possible combinations of the above architectures as the secret and substitute model are tabulated in Table~\ref{tab:grids}.

As is obvious from the table, the agreements along the principal diagonal, i.e.\ corresponding to scenarios where the secret model and substitute model architectures are identical, are in general high. These results also corroborate the findings of \cite{Juuti2018PRADAPA}. We believe that the performance degradation from using a less or more complex substitute model than the secret model results from underfitting or overfitting, respectively. A less complex model may not have the required complexity to fit to the constructed dataset as it is generated by querying a more complex function. Conversely, a more complex model may readily overfit to the constructed dataset, leading to poor generalization and thus a lower agreement score.

Even though the agreements are higher in general for identical complexities, they are still reasonably high even when there is mismatch in model complexities.  Model reverse-engineering can be used to recover information the architecture and hyperparameters of the secret model. Using this information, the adversary can then construct a substitute model that has a similar architecture and a comparable set of hyperparameters, with the hope that the trained substitute model will achieve a better agreement.

%\subsection{Accuracy of the extracted models}

%The accuracy of the extracted model is shown in Table~\ref{tab:accuracy}. It can be seen that with an improvement in agreement, there is a proportionate improvement in accuracy for both image and text classification tasks.

%\begin{table}[t!]
%	\centering
%	\caption{The agreement on the secret test set for text classification tasks. Rows correspond to subset selection strategies, while columns correspond to datasets. The budget is fixed at 10K.}
%	\label{tab:text}	
%	\begin{tabular}{lrrr}
%		\toprule
%		 & IMDB & AG News & QC \\
%		\midrule
%		Random            & 71.67 & 74.51 & 53.00 \\
%		Uncertainty       & 73.48 & 75.47 & \textbf{58.60} \\
%		\textsc{k}-center & \textbf{77.67} & \textbf{75.87} & 56.80 \\
%		\midrule
%		Full thief dataset & 86.38 & 90.07 & 77.80 \\
%		Uniform noise (discrete) & 53.23 & 35.50 & 21.60 \\
%		\bottomrule
%	\end{tabular}
%\end{table}

\subsection{Text classification}
\label{sec:text-results}

%%%%%%%%%%%%%%%%%%%%%%%%%%%%%%%%%%%%%%%%%%%%%%%%%
% START OF TEXT RESULTS
%%%%%%%%%%%%%%%%%%%%%%%%%%%%%%%%%%%%%%%%%%%%%%%%%

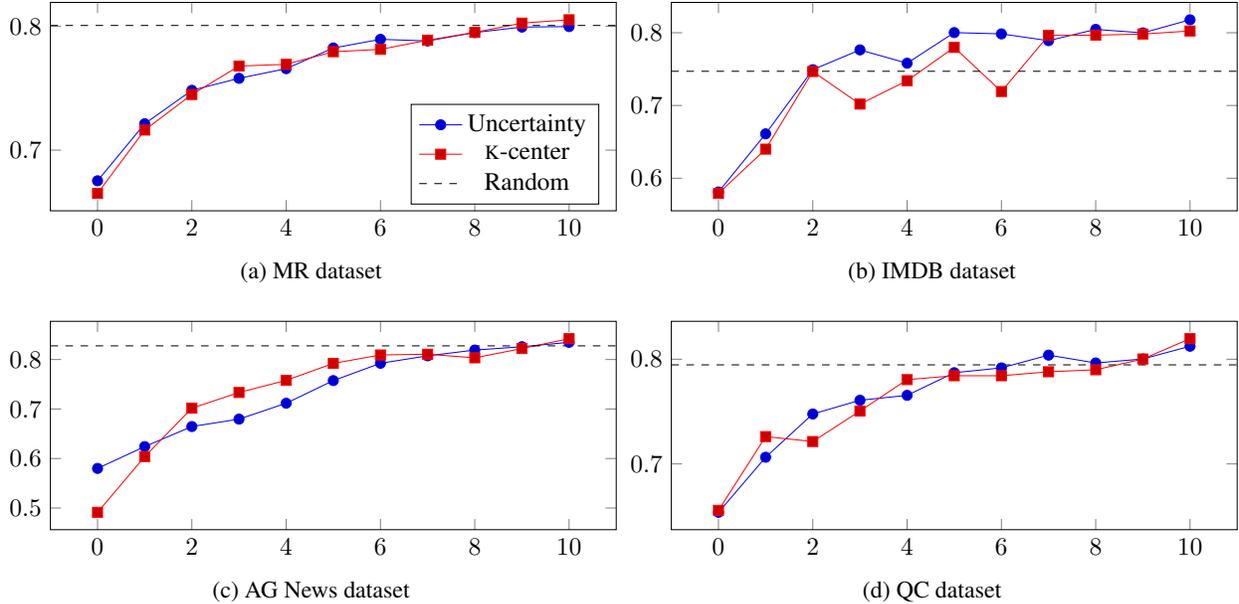
\begin{figure*}[t!]
	\begin{subfigure}[t]{0.5\linewidth}
		\begin{tikzpicture}[scale = 0.95]
		\begin{axis}[legend pos=south east,width=9.5cm,height=4.5cm,ytick={0.7,0.8},xmin=-1,xmax=11]
		\plotfile{mr.dat}
		\legend{Uncertainty,\textsc{k}-center}
		\addplot[mark=none, black, dashed,domain=-1:11] {0.8007};
		\addlegendentry{Random}
		\end{axis}
		\end{tikzpicture}
		\caption{MR dataset}
		\label{tab:iterations-mr}
	\end{subfigure}%
	\begin{subfigure}[t]{0.5\linewidth}
		\begin{tikzpicture}[scale = 0.95]
		\begin{axis}[legend pos=south east,width=9.5cm,height=4.5cm,xmin=-1,xmax=11]
		\plotfile{imdb.dat}
		\legend{Uncertainty,\textsc{k}-center}
		\addplot[mark=none, black, dashed,domain=-1:11] {.747};
		\addlegendentry{Random}
		\legend{}
		\end{axis}
		\end{tikzpicture}
		\caption{IMDB dataset}
		\label{tab:iterations-imdb}
	\end{subfigure}\par\bigskip
	\begin{subfigure}[t]{0.5\linewidth}
		\begin{tikzpicture}[scale = 0.95]
		\begin{axis}[legend pos=south east,width=9.5cm,height=4.5cm,xmin=-1,xmax=11]
		\plotfile{agnews.dat}
		\legend{Uncertainty,\textsc{k}-center}
		\addplot[mark=none, black, dashed,domain=-1:11] {.8276};
		\addlegendentry{Random}
		\legend{}
		\end{axis}
		\end{tikzpicture}
		\caption{AG News dataset}
		\label{tab:iterations-fashionmnist}
	\end{subfigure}%
	\begin{subfigure}[t]{0.5\linewidth}
		\begin{tikzpicture}[scale = 0.95]
		\begin{axis}[legend pos=south east,width=9.5cm,height=4.5cm,ytick={0.7,0.8},xmin=-1,xmax=11]
		\plotfile{qc.dat}
		\legend{Uncertainty,\textsc{k}-center}
		\addplot[mark=none, black, dashed,domain=-1:11] {0.7946};
		\addlegendentry{Random}
		\legend{}
		\end{axis}
		\end{tikzpicture}
		\caption{QC dataset}
		\label{tab:iterations-fashionmnist}
	\end{subfigure}
	\caption{The improvement in agreement for text classification experiments with a total budget of 20K over 10 iterations. Since random is not run iteratively, it is indicated as a line parallel to the X-axis.}
	\label{fig:iterations-text}
\end{figure*}

\begin{table*}[t!]
	\caption{The agreement on the secret test set for text classification tasks. Each row corresponds to a subset selection strategy, while each column corresponds to a query budget.}
	\label{tab:results-text}
	\begin{subfigure}[t]{0.5\linewidth}
		\centering
		\caption{MR dataset}
		\label{tab:datasets-mr}
		\begin{tabular}{lrrrrr}
			\toprule
			& 10K & 15K & 20K & 25K & 30K\\
			\midrule
			Random & 76.45 & 78.24 & 79.46 & 81.33 & 82.36\\
			Uncertainty & \textbf{77.19} & 80.39 & 81.24 & \textbf{84.15} & 83.49\\
			\textsc{k}-center & 77.12 & \textbf{81.24} & \textbf{81.96} & 83.95 & \textbf{83.96}\\
			\midrule
			\multicolumn{5}{l}{Using the full thief dataset (89K):}  & 86.21\\
			\multicolumn{5}{l}{Using discrete uniform noise samples (100K):}  & 75.79\\
			\bottomrule
		\end{tabular}
	\end{subfigure}%
	\begin{subfigure}[t]{0.5\linewidth}	
		\centering
		\caption{IMDB dataset}
		\label{tab:datasets-imdb}
		\begin{tabular}{lrrrrr}
			\toprule
			& 10K & 15K & 20K & 25K & 30K\\
			\midrule
			Random & 71.67 & 78.79 & 74.70 & 80.71 & 79.23\\
			Uncertainty & 73.48 & 78.12 & \textbf{81.78} & \textbf{82.10} & 82.17\\
			\textsc{k}-center & \textbf{77.67} & \textbf{78.96} & 80.24 & 81.58 & \textbf{82.90} \\
			\midrule
			\multicolumn{5}{l}{Using the full thief dataset (89K):}  & 86.38\\
			\multicolumn{5}{l}{Using discrete uniform noise samples (100K):}  & 53.23\\
			\bottomrule
		\end{tabular}
	\end{subfigure}\par\bigskip
	\begin{subfigure}[t]{0.5\linewidth}
		\centering
		\caption{AG News dataset}
		\label{tab:datasets-agnews}
		\begin{tabular}{lrrrrr}
			\toprule
			& 10K & 15K & 20K & 25K & 30K\\
			\midrule
			Random & 74.51 & 80.39 & 82.76 & 83.97 & 84.20\\
			Uncertainty & 75.47 & \textbf{82.08} & 83.47 & 84.96 & \textbf{87.04}\\
			\textsc{k}-center & \textbf{75.87} & 79.63 & \textbf{84.21} & \textbf{84.97} & 85.96\\
			\midrule
			\multicolumn{5}{l}{Using the full thief dataset (89K):}  & 90.07\\
			\multicolumn{5}{l}{Using discrete uniform noise samples (100K):}  & 35.50\\
			\bottomrule
		\end{tabular}
	\end{subfigure}%
	\begin{subfigure}[t]{0.5\linewidth}
		\centering
		\caption{QC dataset}
		\label{tab:datasets-qc}
		\begin{tabular}{lrrrrr}
			\toprule
			& 10K & 15K & 20K & 25K & 30K\\
			\midrule
			Random & 53.00 & 58.00 & 57.20 & 64.40 & 60.40\\
			Uncertainty & \textbf{58.60} & 65.20 & 64.40 & 65.60 & 69.20\\
			\textsc{k}-center & 56.80 & \textbf{65.60} & \textbf{68.60} & \textbf{67.40} & \textbf{71.80}\\
			\midrule
			\multicolumn{5}{l}{Using the full thief dataset (89K):}  & 77.80\\
			\multicolumn{5}{l}{Using discrete uniform noise samples (100K):}  & 21.60\\
			\bottomrule
		\end{tabular}
	\end{subfigure}
\end{table*}

%% END

In addition to the image domain we also present the results of running our framework on datasets from the text domain.
For each text dataset (described in Section~\ref{sec:datasets}), we run our framework across the following total query budgets: 10K, 15K, 20K, 25K and 30K. As it is non-trivial to modify DeepFool to work on text, we omit the strategies that make use of it.

The results of our experiments on the text classification tasks are shown in Table~\ref{tab:results-text}. Like for the image classification tasks, for \todov{a budget of} 20K, we show the agreement at the end of each iteration for every strategy and each dataset in Figure~\ref{fig:iterations-text}.

\paragraph{Effectiveness of active learning\todov{.}} As in the case of images, the use of intelligent selection of thief dataset samples peforms better: there is no dataset for which the random strategy performs better than all of the other strategies. In particular, \textsc{k}-center and uncertainty underperform only once each. Furthermore, incremental improvement over iterations is evident in the case of text, as seen in Figure~\ref{fig:iterations-text}.

\paragraph{\clearcolor{Impact of the query budget}\todov{.}} \todoy{As in the case of images we observe a similar pattern in the text results where there is usually an improvement in agreement when increasing the total query budget.}

\paragraph{Effectiveness of the universal thief\todov{.}} Once again, all 3 experiments using the thief dataset (random included) perform significantly better than the uniform noise baseline. On an average, using the full thief dataset (89K) leads to an improvement in agreement by $\mathbf{2.22}\times$ over the discrete uniform baseline. Even with a budget of 30K, an improvement of $\mathbf{2.11}\times$ is retained with active learning.\\

In summary, our experiments on the text dataset illustrate that the framework is not restricted to image classification, but may be used with tangible benefits for other media types as well.

\section{Related work}
%\todos{\shortciteauthor{DBLP:journals/corr/TramerZJRR16}, \shortciteauthor{DBLP:journals/corr/PapernotMGJCS16} and \shortciteauthor{DBLP:journals/corr/abs-1806-05476}} attempt model extraction through random noise, domain-specific data and public datasets respectively. However, their approach falls short in several key criteria, as shown in Table~\ref{tab:comparison}.
We discuss related work in three broad areas: model extraction, model reverse-engineering and active learning.
\subsection{\clearcolor{Model extraction}}
\noindent\textbf{Attacks.} \shortciteauthor{DBLP:journals/corr/TramerZJRR16} present the first work on model extraction, and the one that is closest to our setting of an adversary with a limited query budget. They introduce several methods for model extraction across different classes of models -- starting with exact analytical solutions (where feasible) to gradient-based approximations (for shallow feedforward neural networks). However, as we demonstrated in Section~\ref{sec:image-results}, their approach of using random uniform noise as a thief dataset for DNNs fails for deeper networks.

\shortciteauthor{7943475} perform model extraction by train a deep learning substitute model to approximate the functionality of a traditional machine learning secret model. In particular, they demonstrate their approach on na{\"i}ve Bayes and SVM secret models trained to perform text classification. They also show that the reverse is not true -- models of lower complexity, viz., na{\"i}ve Bayes or SVM are unable to learn approximations of more complex deep learning models.

\shortciteauthor{2017arXiv170307909S} present a Seed-Explore-Exploit framework whereby an adversary attempts to fool a security mechanism with a ML-based core, e.g., a CAPTCHA system that uses click time to determine whether the user is benign or a bot. They use model extraction to inform the generation of adversarial examples that allows the attacker to perturb inputs to bypass detection. To do this, they use the Seed-Explore-Exploit framework, which starts with a benign and malicious seed, and proceeds by using the Gram-Schmidt process to generate orthonormal samples near the mid-points of any two randomly selected seed points of opposite classes in the exploration phase. These are then used to train a substitute model, which provides useful information in the generation of adversarial examples (during the exploitation phase).

\shortciteauthor{DBLP:journals/corr/abs-1811-02054} draw parallels between model extraction and active learning. They demonstrate that \textit{query synthesis} (QS) active learning can be used to steal ML models such as decision trees by generating queries de novo, independent of the original dataset distribution. They implement two QS active learning algorithms and use them to extract binary classification models ($d$-dimensional halfspaces). In contrast to their approach, ours uses \textit{pool-based} active learning.

\shortciteauthor{8574124} make use of active learning in conjunction with problem domain data to extract a shallow feedforward neural network for text applications, when interfacing with the secret model through APIs with strict rate limits. \shortciteauthor{8642683} design an exploratory attack that uses a generative adversarial network (GAN) trained on a small number of secret dataset samples, which is then able to generate informative samples to query the secret model with. In both these works, the extracted model is then used to launch evasion attacks (i.e.\ finding samples which the secret model incorrectly labels) and causative attacks (i.e.\ exploiting classifiers trained by user feedback by intentionally providing it mislabeled data).

\noindent\textbf{Defenses.} \shortciteauthor{quiring2017fraternal} show that when the secret model is a decision tree, defenses against model watermarking can also be used as defenses for model extraction attacks. This defense is only applicable to decision trees, and does not apply to DNNs.

\shortciteauthor{DBLP:journals/corr/abs-1806-00054} apply a perturbation to the predicted softmax probability scores to dissuade model extraction adversaries. Of course, such a defense would still leave the secret model vulnerable to attacks that can work with only Top-1 predictions to be returned, such as ours. Of course, we speculate that it may lead to a lower agreement in our approach if the adversary does not identify the defense ahead of time and continues to operate on the perturbed softmax outputs directly.

\shortciteauthor{Juuti2018PRADAPA} design PRADA, a framework to detect model extraction attacks by computing the empirical distribution of pairwise distances between samples. They demonstrate that for natural samples (i.e.\ benign inputs to an MLaaS API), the distribution of pairwise distances is expected to fit a bell curve, whereas for noisy samples a peaky distribution is observed instead. The queries made by a client can be logged and the distribution can be analyzed to detect a potential model extraction attack. We speculate that our approach will break this defense, as the universal thief datasets that it pulls from -- while not from the same domain -- are indeed otherwise natural, and we expect pairwise distances between samples to fit a bell curve.

\shortciteauthor{DBLP:journals/corr/abs-1808-00590} design MLCapsule, a guarded offline deployment of MLaaS using Intel SGX. This allows providers of machine learning services to serve offline models with the same security guarantees that are possible for a custom server-side deployment, while having the additional benefit that the user does not have to trust the service provider with their input data. They demonstrate an implementation of PRADA \cite{Juuti2018PRADAPA} within MLCapsule as a defense against model extraction attacks.

\shortciteauthor{xu2018deepobfuscation} obfuscate CNN models by replacing the complex CNN feature extractors with shallow, sequential convolution blocks. Networks with 10s or 100s of layers are simulated with a shallow network with 5-7 convolution layers. The obfuscated secret model is shown to be more resilient to both \textit{structure piracy} (i.e.\ model reverse-engineering) and \textit{parameter piracy}, thus dissuading model extraction attackers. We speculate that our approach will still be able to extract the model if it is given access to the obfuscated model through the same API interface.

\shortciteauthor{kesarwani2018model} design a \textit{model extraction monitor} that logs the queries made by users of a MLaaS service. They use two metrics -- total information gain and coverage of the input feature space by the user's queries -- in order to detect a possible model extraction attack, while minimizing computational overhead. They demonstrate their monitor for decision tree and neural network secret models. We speculate that our approach may be detected by such a model extraction monitor, however an informed adversary could choose to tweak the active learning subset selection strategy to avoid detection by picking samples with lower information gain, and covering only a limited portion of the feature space.

\noindent\textbf{Applications.} \shortciteauthor{DBLP:journals/corr/PapernotMGJCS16} use model extraction for the generation of adversarial examples. They query a limited subset of the training data, or hand-crafted samples that resemble it, against the secret model. The resulting labels are then used to train a crude substitute model with a low test agreement. A white-box adversarial example generation technique is used to generate adversarial examples, which are then used to attack the original secret model by leveraging the transferability of adversarial examples.

\subsection{Model reverse-engineering}
As we show in Section~\ref{sec:grids}, while the agreement obtained by us is respectable even when the secret model and substitute model architectures do not match, agreement is improved when they match. Thus, it is in the best interest of the adversary to try to obtain information about the secret model architecture -- this is possible through model reverse-engineering.

\shortciteauthor{2017arXiv171101768O} train a meta-model which takes as input the softmax prediction probabilities returned by the secret model and predicts, with statistically significant confidence, secret model hyperparameters such as the number of convolution layers, the filter size of CNNs, the activation function used, the amount of dropout, batch size, optimizer, etc. To do this, they first randomly generate and trains networks of varying complexity and queries them to create a dataset to train the meta-model on.

\shortciteauthor{DBLP:journals/corr/abs-1802-05351} estimate the regularizer scale factor $\lambda$ for linear regression (ridge regression and LASSO), kernel regression (kernel ridge regression), linear classification (SVM with hinge loss, SVM with squared hinge loss, $L_1$-regularized logistic regression and $L_2$-regularized logistic regression) and kernel classification algorithms (kernel SVM with hinge loss, kernel SVM with squared hinge loss).

\shortciteauthor{DBLP:journals/corr/abs-1812-11720} use a timing side channel for model reverse-engineering, i.e.\ they use the execution time of the forward pass of the secret model, averaged across queries, to infer model architecture and hyperparameters. This information is then used to reduce the search space by querying a pretrained regressor trained to map execution time to hyperparameters (such as the number of layers). Further search is performed using a reinforcement learning algorithm that predicts the best model architecture and hyperparameters in this restricted search space.

\shortciteauthor{yan2018cache} use a similar insight that the forward pass of DNNs rely on GeMM (generalized matrix multiply) library operations. They use information from cache side channels to reverse engineer information about DNN architectures, such as the number of layers (for a fully connected network) and number of filters (for a CNN). However, such an attack cannot determine the presence and configuration of parameter-free layers such as activation and pooling layers.

\shortciteauthor{hong2018security} present another attack using cache side channels that monitors the shared instruction cache. The attacker periodically flushes the cache lines used by the victim secret model and measures access time to the target instructions. This side channel information is then used to reconstruct the architecture, including parameter-free layers.

\shortciteauthor{hu2019neural} use bus snooping techniques (passively monitoring PCIe and memory bus events). Using this information, they first infer kernel features such as read and write data volume of memory requests. This information is then used to reconstruct the layer topology and predict the network architecture.

\subsection{Active learning}
There is an existing body of work on active learning, applied traditionally to classic machine learning models such as na{\"i}ve Bayes and SVMs. We refer the reader to the survey by \shortciteauthor{settles2009active} for details.

Active learning methods engineered specifically for deep neural networks include the following:
\begin{itemize}
\item \shortciteauthor{sener2017active} present an active learning strategy based on core-set construction. The construction of core-sets for CNNs is approximated by solving a \textsc{k}-center problem. The solution is further made robust by solving a mixed integer program that ensures the number of outliers does not exceed a threshold. They demonstrate significant improvements, when training deep CNNs, over earlier active learning strategies (such as uncertainty) and over a \textsc{k}-median baseline.
\item \shortciteauthor{DBLP:journals/corr/abs-1802-09841} present a margin-based approach to active learning. The DeepFool \cite{moosavi2016deepfool} method for generation of adversarial examples is used to generate samples close to the decision boundary by perturbing an input image until the class predicted by the image classification model changes. They demonstrate that their method is competitive to that of \cite{sener2017active} for image classification tasks on CNNs, while significantly outperforming classical methods (such as uncertainty).
\end{itemize}

\section{Conclusion}
In this paper, we introduce three criteria for practical model extraction. Our primary contribution is a novel framework that makes careful use of unlabeled public data and active learning to satisfy these criteria.
%We evaluate our framework across a number of datasets, with encouraging results.

We demonstrate the effectiveness of our framework by successfully applying it to a diverse set of datasets. Our framework is able to extract DNNs with high test agreement and on a limited query budget, using only a fraction (10-30\%) of the data available to it.
%Although we restrict our experimentation to image and text classification, our technique is not task-specific, and should generalize to other supervised learning tasks as well.

Future work on developing this method of attack includes the development of better active learning strategies and the exploration of other novel combinations of existing active learning strategies.

\section*{Acknowledgement}

We would like to thank Somesh Jha for his helpful inputs. We thank NVIDIA for providing us computational resources, and Sonata Software Ltd.\ for partially funding this work.

\bibliographystyle{unsrtnat}  
\bibliography{references}  %%% Remove comment to use the external .bib file (using bibtex).

\end{document}